\documentclass[12pt,reqno]{article}
\usepackage{graphicx}
\usepackage{cite}
\usepackage{amsmath}
\usepackage{amssymb}
\usepackage{subcaption}
\usepackage{hyperref}
\usepackage{geometry} 
\graphicspath{{{./}}}

\usepackage{xcolor}
\newcommand{\ADD}[1]{{#1}}

\newcommand{\aref}[1]{App.\,\ref{#1}}
\newcommand{\fref}[1]{Fig.\,\ref{#1}}
\newcommand{\tref}[1]{Table\,\ref{#1}}
\newcommand{\eref}[1]{Eq.\,(\ref{#1})}

\newcommand{\sref}[1]{Sec.\!~\ref{#1}}

\newcommand{\cref}[1]{Ref.\,\cite{#1}}
\newcommand{\crefs}[1]{Refs.\,\cite{#1}}

\newcommand{\vs}{{\it vs.}\! }
\newcommand{\ie}{{\it i.e.}\!\, }
\newcommand{\eg}{{\it e.g.}\!\, }

\newcommand{\etal}{{\it et al.} }

\newcommand{\Cbb}{\mathbb{C}}

\newcommand{\eb}{\mathbf{e}}

\newcommand{\pb}{\mathbf{p}}

\newcommand{\nb}{\mathbf{n}}
\renewcommand{\sb}{\mathbf{s}}
\newcommand{\Ab}{\mathbf{A}}

\newcommand{\Eb}{\mathbf{E}}

\newcommand{\Qb}{\mathbf{Q}}

\newcommand{\Lb}{\mathbf{L}}

\newcommand{\Ib}{\mathbf{I}}
\newcommand{\Rb}{\mathbf{R}}
\newcommand{\Sb}{\mathbf{S}}

\newcommand{\Xb}{\mathbf{X}}

\newcommand{\As}{\mathsf{A}}
\newcommand{\Ds}{\mathsf{D}}

\newcommand{\xs}{\mathsf{x}}
\newcommand{\ys}{\mathsf{y}}
\newcommand{\Ws}{\mathsf{W}}
\newcommand{\bs}{\mathsf{b}}

\newcommand{\Is}{\mathsf{I}}
\newcommand{\Us}{\mathsf{U}}
\newcommand{\Ls}{\mathsf{L}}
\newcommand{\Qs}{\mathsf{Q}}
\newcommand{\epsilonb}{{\boldsymbol{\epsilon}}}
\newcommand{\thetab}{{\boldsymbol{\theta}}}
\newcommand{\sigmab}{{\boldsymbol{\sigma}}}
\newcommand{\phib}{{\boldsymbol{\phi}}}
\newcommand{\Lambdab}{{\boldsymbol{\Lambda}}}

\newcommand{\tr}{{\operatorname{tr}}}

\newcommand{\diag}{{\operatorname{diag}}}

\newcommand{\DGCNN}{{dGCNN}}
\newcommand{\RGCNN}{{rGCNN}}
\newcommand{\arch}[3]{({#1}\text{:}{#2}\text{:}{#3})}

\newcommand{\error}{\varepsilon}
\newcommand{\corr}{C}

\newcommand{\Nf}{N_\text{f}}
\newcommand{\Nc}{N_\text{c}}
\newcommand{\Nd}{N_\text{d}}

\title{\bf
\ADD{Mesh-based graph convolutional neural networks for modeling materials with microstructure}
}
\author{Ari Frankel, Cosmin Safta, Coleman Alleman, Reese Jones\footnote{Corresponding author: \tt rjones@sandia.gov}\\
{\it Sandia National Laboratories, Livermore, CA 94551}
}

\begin{document}
\date{}

\maketitle{}

\begin{abstract}
Predicting the evolution of a representative sample of a material with microstructure is a fundamental problem in homogenization.
In this work we propose a graph convolutional neural network that utilizes the discretized representation of the initial microstructure directly, without segmentation or clustering.
Compared to feature-based and pixel-based convolutional neural network models, the proposed method has a number of advantages:
(a) it is deep in that it does not require featurization but can benefit from it,
(b) it has a simple implementation with standard convolutional filters and layers,
(c) it works natively on unstructured and structured grid data without interpolation (unlike pixel-based convolutional neural networks),
and
(d) it preserves rotational invariance like other graph-based convolutional neural networks.
We demonstrate the performance of the proposed network and compare it to traditional pixel-based convolution neural network models and feature-based graph convolutional neural networks on \ADD{multiple} large datasets.
\end{abstract}

\section{Introduction}
Predicting the evolution of a system with a complex initial state represents a wide class of physical problems of scientific and technological interest.
For instance, simulating the evolution of materials with complex microstructure is necessary for predicting the behavior of highly engineered materials \cite{kraft1996extended,yin2008statistical,ghosh2011computational,stenzel2016predicting,li2017review,herriott2020predicting}.
With the advent of machine learning for physical applications and the availability of considerable experimental and high-fidelity simulation data, models and architectures for these and related applications have begun to arise \cite{frankel2019oligocrystals,frankel2019evolution,vlassis2020geometric,pandey2020machine}.
These models can be used for a number of tasks such as sub-grid accurate constitutive modeling \cite{frankel2019oligocrystals}, material design by structure property exploration \cite{noh2019inverse}, and uncertainty quantification of materials with high intrinsic variability \cite{khalil2021modeling}.
For this work, we are interested in predicting the evolution of the physical response of a sample given its initial state and a history of loading.
For this class of problems, we assume the initial state can be represented as a field or collection of fields captured in a multispectral/multichannel image and this image is data on a structured grid or an unstructured mesh.

In the ever expanding field of machine learning (ML) \cite{bishop2006pattern,hastie2005elements,goodfellow2016deep}, there are many methods suitable to the task of supervised learning where the objective is to represent an input-output map to high fidelity.
Neural networks (NN) \cite{hopfield1982neural,Goodfellow-et-al-2016} are a particularly versatile sub-category of machine learning techniques suitable for regression tasks.
They can be designed to be smooth, expressive models of physical behavior and have been shown to be effective in reducing complex processes to low-dimensional latent spaces \cite{lee2019deep,fulton2019latent}.
In particular, convolutional neural networks (CNNs) \cite{lecun1989backpropagation,o2015introduction,albawi2017understanding,qin2018convolutional} are an efficient version of fully connected networks applied to image data.
Their main advantage is the reduction of the weight space needed to encode images to a manageable size by exploiting spatial correlation, locality, and similarity with a compact kernel filter.
They have been spectacularly successful in a variety of image (and time series) applications \cite{lecun1998gradient,krizhevsky2012imagenet,liu2017survey}.

Graph convolutional neural networks (GCNNs) \cite{bruna2013spectral,bronstein2017geometric} can be seen as counterparts to CNNs which operate on topologically related, as opposed to spatially or temporally related, data.
Bruna \etal \cite{bruna2013spectral} recognized that many of the properties that make CNNs effective on gridded data could be translated to graph-based data, such as data on unstructured meshes.
They developed both local (\eg pooling) and spectral/global (\eg convolution in the Fourier domain) operations on graph data.
They made the connection with Fourier bases through the graph Laplacian (of a binary adjacency matrix) to translate the convolution operation to graphs.
In another arena (graph wavelets/graph signal processing), some of the graph convolution developments were preceded by Hammond \etal \cite{hammond2011wavelets} who developed the mathematics of spectral analysis and filtering on the more general context of kernel-weighted graphs.
In order to surmount the global and expensive nature of applying filters in the spectral domain, where an eigen-decomposition is required, Defferrard \etal \cite{defferrard2016convolutional} (ChebNet) introduced (spatially) compact polynomial filters.
In particular, they approximated the action of a general spectral filter with a truncated  Chebychev expansion of the eigenvalue matrix of the graph Laplacian,
leveraging the fact that repeated application of these filters ($k$ times) leads to $k$-hop diffusion of information.
Kipf and Welling \cite{kipf2016semi} took these developments to their logical conclusion with the Graph Convolutional Network (GCN).
They truncated the Chebychev expansion to first order and relied on deep/multi-layer networks to build expressive representations.\footnote{\aref{app:gcn} provides a brief synopsis of the GCN which we base our developments on.}
For a more complete overview of GCNNs see reviews by Wu \etal \cite{wu2020comprehensive}. Zhang \etal \cite{zhang2020deep}, and  Zhou \etal \cite{zhou2020graph}.
Note that related kernel-based NN methods, such as the work of Trask \etal \cite{trask2019gmls,trask2020enforcing} exists and have been shown to be effective in physical applications.
Also, Bronstein \etal \cite{bronstein2017geometric} provides an insightful perspective on applications of deep learning to non-Euclidean data, graphs, and manifolds as well as the relevant mathematics.

One of the issues with using convolutional NNs for physical problems is the need to preserve fundamental physical relationships such as frame indifference, where scalars are unchanged (invariant) to rotational changes in observer and tensorial quantities rotate in a corresponding way (equivariance) with rotation of the observation coordinate frame.
The advantages of preserving these spatial symmetries in filter based NNs was recognized early on \cite{rowley1998rotation}.
By construction CNNs embed shift/translation invariance where every local neighborhood is processed in a similar manner.
Traditional convolutional filters operate strictly on structured grids of pixelated images and generally do not preserve rotational symmetries but provide a richer set of filters than methods that do.
Numerous efforts have been made to endow CNNs with rotational invariance.
Dielemann \etal \cite{dieleman2016exploiting} augmented the layers of a CNN to add rotations of the image by $\pi/4$ and inversions.
Cohen and Welling \cite{cohen2016group} outlined the mathematics whereby convolutional network will be equivariant with respect to any group, including rotation and reflections.
Worrall \etal \cite{worrall2017harmonic}  developed filters based on circular harmonics.
Chidester \etal \cite{chidester2018rotation} used a (discrete) Fourier transform to embed rotational invariance by filter augmentation.
Using concepts from abstract algebra, Kondor and Trivedi \cite{kondor2018generalization} proved that convolutional structure is a necessary and sufficient condition for equivariance to the action of a compact symmetry group.
This finding serves as a requirement for CNN to be equivariant.
Recently, Finzi \etal \cite{finzi2020generalizing} provided a significant extension of Cohen and Welling treatment of small, discrete symmetry groups to continuous (Lie) groups, \eg rotations in 3D (the special orthogonal group SO(3)).
In contrast to pixel-based CNNs, the existing graph based filters can operate on unstructured spatial data by use of user-defined neighbor attributions but lose some spatial information in the process.
Furthermore, graph-based convolutional networks can have an inherent rotational invariance (assuming the node data is invariant/equivariant) since the representation has been lifted out of its spatial embedding.

Currently, there are many applications of \ADD{pixel}-based CNNs to physics and mechanics problems \ADD{that have data on a structured grid}.
Some of the early work focused on classification and simple property prediction.
Chowdhury \etal \cite{chowdhury2016image} used a CNN to classify microstructures.
Lubbers \etal \cite{lubbers2017inferring} developed a CNN based method for inferring low-dimensional microstructure representations and used the model for microstructure generation.
DeCost \etal \cite{decost2017exploring} also applied CNNs to microstructure representations and used the model to connect processing to structure.
Contemporaneous with these developments, Kondo \etal \cite{kondo2017microstructure} used a CNN to predict ionic conductivity based on microstructure.
Many publications have followed these initial applications, some have targeted evolution prediction tasks.
Frankel and collaborators devised a hybrid between a CNN and recurrent NN (RNN) network to predict the elastic-plastic stress response of polycrystalline samples~\cite{frankel2019oligocrystals} and have used a convolutional long short-term memory architecture (convLSTM)~\cite{xingjian2015convolutional}, which integrates CNN and RNN  aspects into a single architecture, to predict the evolution of the stress field~\cite{frankel2019evolution}.
The latter work~\cite{frankel2019evolution} was based on convLSTM network of Shi \etal~\cite{shi2015convolutional} which was developed for atmospheric predictions and provides a framework for representing the solutions of time dependent partial differential equations.
On the other hand, we are aware of only a few applications of GCNNs to physics or mechanics to date.
Vlassis \etal \cite{vlassis2020geometric} employed a feature-based graph neural network to model the elastic interaction of grains in polycrystalline samples using an adaptation of the CNN-RNN network in \cref{frankel2019oligocrystals}.
Chen \etal \cite{chen2020estimating} employed a GCNN trained to  sparse (and unstructured) diffusion data to model human tissue.

In contrast to these approaches, the proposed graph based convolutional neural network processes the structured or unstructured microstructural images directly and in an invariant manner.
The adjacency matrix, which conveys/defines neighbors and spatial information, is based on the topology of the image data itself.
The formulation does not require an obvious segmentation of the microstructure, which may be occluded by noise in real data.
Most importantly it does not require featurization of the multi-channel/hyperspectral image data \cite{bessa2017framework,mozaffar2019deep}.
Nevertheless, as we will show, it can benefit from obvious features but no feature engineering is needed to obtain good accuracy.
The main contributions of the work are: a generalization of graph/pixel CNNs for predicting homogenized response of samples with complex microstructure, and a demonstrative comparison of performance relative to existing methods.

In \sref{sec:problem} we describe the physical problem which is a homogenization of physical response suitable for sub-grid/multi-scale applications~\cite{jones2018machine,frankel2019oligocrystals}.
In particular we apply the methods to homogenization of the evolution of stress of a sample volume where internal state determined by microstructure.
In \sref{sec:architecture} we describe the proposed neural network architecture and relate it to traditional CNNs and feature-engineered GCNNs.
In \sref{sec:results} we focus on comparing the performance of pixel-based convolutional neural network (CNN), a feature-based ``reduced'' graph convolutional neural network (\RGCNN), and the proposed ``direct'' graph-based convolutional neural network (\DGCNN) \ADD{which operates directly on structured or unstructured image data like a CNN without the need for featurization or segmentation required by the \RGCNN}.
\ADD{
To assess the performance of these models, we employ two exemplars of the homogenization problem: (a) the prediction of the stress evolution in a porous metal and (b) the prediction of the stress evolution in a polycrystalline material.
With these physical problems we created four datasets using: (a) an ensemble of three-dimensional (3D) realizations of the porous metal,  (b) an ensemble of two-dimensional (2D) realizations of polycrystals, (c) a 3D ensemble with low variance in the grain sizes, and (d) a 3D ensemble with high variance in the grain sizes, all of which are subjected to a tension deformation process.
The first dataset is used to demonstrate the efficacy of the proposed mesh-based GCNNs on unstructured meshes; the remainder are used to compare CNNs to GCNNs on structured meshes.
We exploit the lower computational cost of the 2D polycrystalline dataset in a deeper exploration of variants and hyper-parameters than would be possible with the 3D data.
}
In \sref{sec:discussion} we summarize the findings of exploring data and parameter efficiencies, architecture variations, and adjacency manipulation, and discuss ongoing work.

\section{Physical problem} \label{sec:problem}
Predicting the physical response of a sample given a complex initial state is representative of a general class of problems in homogenization \cite{mura2013micromechanics,nemat2013micromechanics}.
In particular, we focus on the prediction of the evolution of the volume average of stress $\sigmab$ in a representative volume $V$ :
\begin{equation} \label{eq:problem}
\bar{\sigmab}(t) = \frac{1}{V} \int \sigmab\left( \epsilonb(t), \phib(\Xb) \right) \, \mathrm{d}\Xb \ ,
\end{equation}
given a microstructural field $\phib(\Xb)$ observed at time $t=0$ and a time-dependent, homogeneous loading determined by the imposed strain $\epsilonb(t)$.
This class of problems is the basis for multiscale models \cite{trovalusci2009genesis}, material structure optimization \cite{le2012material}, and material variability and uncertainty quantification \cite{khalil2021modeling}.
The microstructural field $\phib(\Xb)$ characterizes location-dependent inhomogeneity that influences the state of the material,  where $\Xb$ is the position vector in the reference configuration of the sample.
Examples of $\phib$ include phase in a multiphase composites \cite{bouquerel2006microstructure}, elastic modulus in a material with inclusions \cite{roduit2009stiffness} or pores \cite{heckman2020automated}, and the local defect density in a defective material \cite{vcekada2007sem}.

Rotational equivariance requires
\begin{equation} \label{eq:invariance}
\Qb \boxtimes \bar{\sigmab}(t,\phib) = \frac{1}{V} \int \sigmab\left( \Qb \boxtimes \epsilonb(t), \Qb \boxtimes \phib(\Xb) \right) \, \mathrm{d}\Xb \ ,
\end{equation}
where $\Qb$ is an orthogonal tensor (rotation) and $\boxtimes$ is the Kronecker product, which is defined as $\Qb \boxtimes \Ab = \Qb \Ab \Qb^T = \sum_{i,j} A_{ij} \Qb \eb_i \otimes \Qb \eb_j$  for a second order tensor $\Ab$.
In effect this is a requirement that rotation of the inputs by $\Qb$ must lead to a corresponding rotation of the output by $\Qb$.
It has fundamental consequences for the form that the function $\sigmab(\epsilonb,\phib)$ is allowed to take \cite{silhavy2013mechanics,jones2018machine}.

\subsection{\ADD{Exemplars}}\label{sec:exemplar}

\ADD{
We employ two exemplars to test and demonstrate performance of the NN architectures: (a) a porous metal and (b) a polycrystalline metal.
The mechanical response in both systems is complex.
They undergo an elastic-to-plastic transition with loading and heterogeneous deformation due to the microstructure.
For simplicity and data storage/memory considerations we focus on the primary component, $\sigma(t)$, of the volume averaged stress $\bar{\sigmab}$ for both exemplars.
Each dataset was created with standard, well-documented software packages \cite{dream3d,albany,sierra}.
}

\subsubsection{\ADD{Porous plasticity}}\label{sec:porous_plasticity}

\ADD{
In the porous metal exemplar, $\phib(\Xb)$ is the local density field with $\phib(\Xb) = 0$ in the pores and equal to the density of the metal elsewhere, which we normalize to one.
\cref{rizzi2018bayesian} describes a similar material model.
Here aluminum was chosen as a representative material.
}

\ADD{
The metal response follows from a widely-employed $J_2$ elastic-plastic model \cite{lubliner2008plasticity} where the stress $\Sb$ is given by a linear elastic rule:
\begin{equation} \label{eq:stress}
\Sb = \Cbb : \Eb_e
\end{equation}
Here ``:'' is a double inner product that allows the 4th order elastic modulus tensor $\Cbb$ to map the elastic strain $\Eb_e$ to the stress $\Sb$.
Note that $\Eb_e$ is distinct from the applied strain $\epsilonb(t)$ driving the evolution of the sample.
For an isotropic material like aluminum the components of $\Cbb$ reduce to
\begin{equation}
[ \Cbb ]_{ijkl} = \frac{E}{(1+\nu)} \left( \frac{\nu}{(1-2\nu)} \delta_{ij}\delta_{kl} + \frac{1}{2} (\delta_{ik}\delta_{jl} + \delta_{il}\delta_{jk}) \right)
\end{equation}
which depends only on Young's modulus $E=$ 59.2 GPa and Poisson's ratio $\nu=$ 0.33.
}

\ADD{
The plastic flow is derived from the von Mises yield condition
\begin{equation}
\sigma_\text{vm}(\Sb) - \check{\sigma}(\epsilon_p) \le 0
\end{equation}
which limits the elastic regime to a convex region in stress space and offsets the elastic strain $\Eb_e$ from the total strain.
Here $\sigma_\text{vm} = \sqrt{\frac{3}{2} \sb\cdot\sb}$ is the von Mises stress where $\sb = \Sb - \tr(\Sb) \Ib$ is the deviatoric part of $\Sb$, and $\epsilon_p$ is the equivalent plastic which is a measure of the accumulated plastic strain computed from the plastic velocity gradient $\Lb_p$.
The yield limit $\check{\sigma}$ is given by a Voce hardening law
\begin{equation}
\check{\sigma} = Y - H \exp(-\alpha \epsilon_p)
\end{equation}
with parameters: initial yield $Y=$ 200.0 MPa, hardening $H=$ 163.6 MPa, and saturation exponent $\alpha=$ 73.3.
}

\ADD{
Realizations were created by a random placement scheme of spherical voids in the sample cube  with constraints on pore overlap with other pores and the sample boundary \cite{brown2018multiscale}.
This process created unit cells with mean porosity 0.09$\pm$0.03 following a beta distribution and at most 20 pores per cell.
The cubic samples were on the order of 1.5$^3$ mm$^3$  with pore radius $\approx$ 150 $\mu$m (refer to \fref{fig:pore_stress}, which will be discussed in \sref{sec:response}).
Pores in each of the 1121 realizations we created  were explicitly meshed and resulted in unstructured discretizations with 14,640 to 101,360 elements.
}

\ADD{
Each realization was subjected to quasi-static uniaxial tension up to 20\% engineering strain with minimal Dirichlet boundary conditions (no lateral constraints, uniform displacement on the ends).
These simulations were performed with Sierra \cite{sierra}.
From these simulations we extracted microstructure $\phib(\Xb)$, applied strain $\epsilon(t)$, volume-averaged stress $\bar{\sigmab}(t)$  data to demonstrate the efficacy of mesh-based GCNNs in the Results section.
}

\subsubsection{Crystal plasticity}\label{sec:crystal_plasticity}

Our second exemplar of the homogenization problem, \eref{eq:problem}, used a crystal plasticity (CP) constitutive model  where $\phib(\Xb)$ is a field of crystal orientations associated with grains.
Although each grain has a relatively simple response, the collective behavior is difficult to predict without a detailed simulation since each grain influences its neighbors \cite{frankel2019oligocrystals}.
For this exemplar, steel was chosen as a representative material.

The response of each crystal follows an elastic-viscoplastic constitutive relation based on well-known meso-scale models \cite{taylor1934mechanism,kroner1961plastic,bishop1951xlvi,bishop1951cxxviii,mandel1965generalisation,dawson2000computational,roters2010overview}.
For the crystal elasticity, we employed the same linear stress model, \eref{eq:stress}, as in the porous metal exemplar albeit with a different elastic modulus tensor $\Cbb$.
In this case ferrous (face centered) cubic symmetry for $\Cbb$ was assumed, and the independent components of the elastic modulus tensor $\Cbb$ were those of steel: $C_{11}, C_{12}, C_{44} = \{204.6, 137.7, 126.2\}$ GPa.
The overall response reflects the anisotropy of each grain; however, the response of polycrystals with random orientations $\phi(\Xb)$ was determined by the collective response (which tends to isotropy with large sample sizes).
In each crystal, plastic flow
\begin{equation}
\Lb_p = \sum_\alpha \dot{\gamma}_{\alpha} \sb_\alpha \otimes \nb_\alpha
\end{equation}
can occur on any of the 12 face-centered cubic slip planes, where $\Lb_p$ is the plastic velocity gradient, $\dot{\gamma}_{\alpha}$ is the slip rate, $\sb_\alpha$ is the slip direction, and $\nb_\alpha$ is the slip plane normal.
We employed a common power-law form for the slip rate relation
\begin{equation} \label{eq:dot_gamma}
\dot{\gamma}_{\alpha}=\dot{\gamma}_0\left|\frac{\tau_{\alpha}}{g_{\alpha}}\right|^{m-1}\tau_{\alpha} \ ,
\end{equation}
driven by the shear stress $\tau_\alpha$ resolved on slip system $\alpha$.
The reference slip rate was chosen to be $\dot{\gamma}_0$ = 1.0 s$^{-1}$, the rate sensitivity exponent was $m = 20$, \ADD{the initial slip is set to zero}, and the slip resistance $g_{\alpha}$ was given the initial value $g_{\alpha}$ = 122.0 MPa \cite{jones2018machine}.
The slip resistance evolved according to \cite{Kocks1976, mecking1976hardening}
\begin{equation} \label{eq:dot_g}
\dot{g}_\alpha = (H-R_d g_\alpha) \sum_\alpha |\dot{\gamma}_\alpha|
\end{equation}
where the hardening modulus was chosen to be $H = 355.0$ MPa and the recovery constant was $R_d = 2.9$.
\ADD{Both \eref{eq:dot_gamma} and \eref{eq:dot_g} are integrated with standard implicit numerical integrators as part of a global equilibrium solution algorithm.
See \cref{jones2018machine} for additional details.
}

For this exemplar, we created multiple sets of $\{ \phib(\Xb), \epsilon(t); \sigma(t) \}$ data to train and compare the NN models described in the next section:
(a) a 2D dataset consisting of 12,000 realizations \cite{frankel2019oligocrystals},
(b) a 3D dataset consisting of 10,000 realizations with low variance in the grain sizes, and
(c) a corresponding 10,000 realization 3D dataset with high variance in the number of grains per realization.
\ADD{
The nominal sample length for each realization was 1 $\mu$m (a realization is shown in \fref{fig:stress_field}a, which will be discussed in the next section) .
}
The variance in the grain sizes is directly related to the variety of grain topologies in the particular ensemble, this aspect will be used in explorations described in the Results section.

\ADD{
Although the GCNN method can be applied to unstructured grids and complex geometries of different sizes, the CNN cannot without interpolation or some other intermediate data processing.
In this exemplar we chose structured computational grids to facilitate direct comparison of pixel and graph based  methods.
The 2D dataset was computed on a 32$\times$32 structured FE mesh and output over 31 time steps (max strain 0.3\%) ; and
the two 3D CP datasets used a 25$\times$25$\times$25 mesh and output over 51 steps (max strain 0.4\%).
Each polycrystal realization was subjected to quasi-static uniaxial tension at a constant engineering strain-rate of $\dot{\epsilon} = 1$ s$^{-1}$ with minimal Dirichlet boundary conditions.
The time evolution of each system was observed over a limited number of time-steps that covered the physics of interest: the transition from elastic to full plastic flow.
}

Realizations of the microstructure $\phib(\Xb)$ consisted of a crystal orientation vector field that encodes the rotation of a crystal in a reference orientation to that in the polycrystal \cite{frankel2019oligocrystals}.
The orientation vector $\phib$ is the unit eigenvector $\pb$ of the rotation tensor $\Rb$, which takes $\Cbb$ from a canonical orientation to that of a particular grain $\Rb \boxtimes \Cbb$, scaled by the rotation angle $\theta$ around that axis
\begin{equation}
\phib = \theta \pb \ \ \text{such that} \ \ \Rb \pb = \pb \ \text{and} \ \| \pb \| = 1 ,
\end{equation}
where $\theta$ can be obtained from the non-unitary eigenvalues of $\Rb$.
Refer to \cref{frankel2019oligocrystals} for more details.
The computational cell for each realization is a cube, as can be seen in \fref{fig:stress_field}a, which is partitioned into sub-regions, called grains, with distinct $\phib(\Xb)$.
The sub-regions evoke a natural topology for a grain-based graph \cite{vlassis2020geometric}.
All polycrystal realizations where created with Dream3D \cite{dream3d} using spatial correlations to obtain a reasonable number of grains and angle distribution functions that gave a uniform texture.
The 2D simulations were run with Albany \cite{albany} and 3D simulations were run with Sierra \cite{sierra}.
\ADD{
See \crefs{jones2018machine,frankel2019predicting,frankel2020prediction} for related efforts.
}

\subsection{System characterization and response}\label{sec:response}

\ADD{
In each exemplar, the non-linear plasticity model and heterogeneous microstructure evoke a complex response to loading $\epsilonb(t)$.
The local stress fields reflect internal inhomogeneities at the pores or the grain boundaries, as characterized by $\phib(\Xb)$, and display large gradients at elastic-plastic transitions.
Spatial averaging to extract the system response $\bar{\sigmab}(t)$ does some smoothing of the evolution but the range of microstructures evokes a distribution of responses.
The behavior is generally similar across the ensemble of realizations but variations are difficult to predict from simple statistics such as mean grain or pore size.
}

\subsubsection{Porous plasticity} \label{sec:pp_response}

\begin{figure}
\centering
{\includegraphics[width=0.55\textwidth]{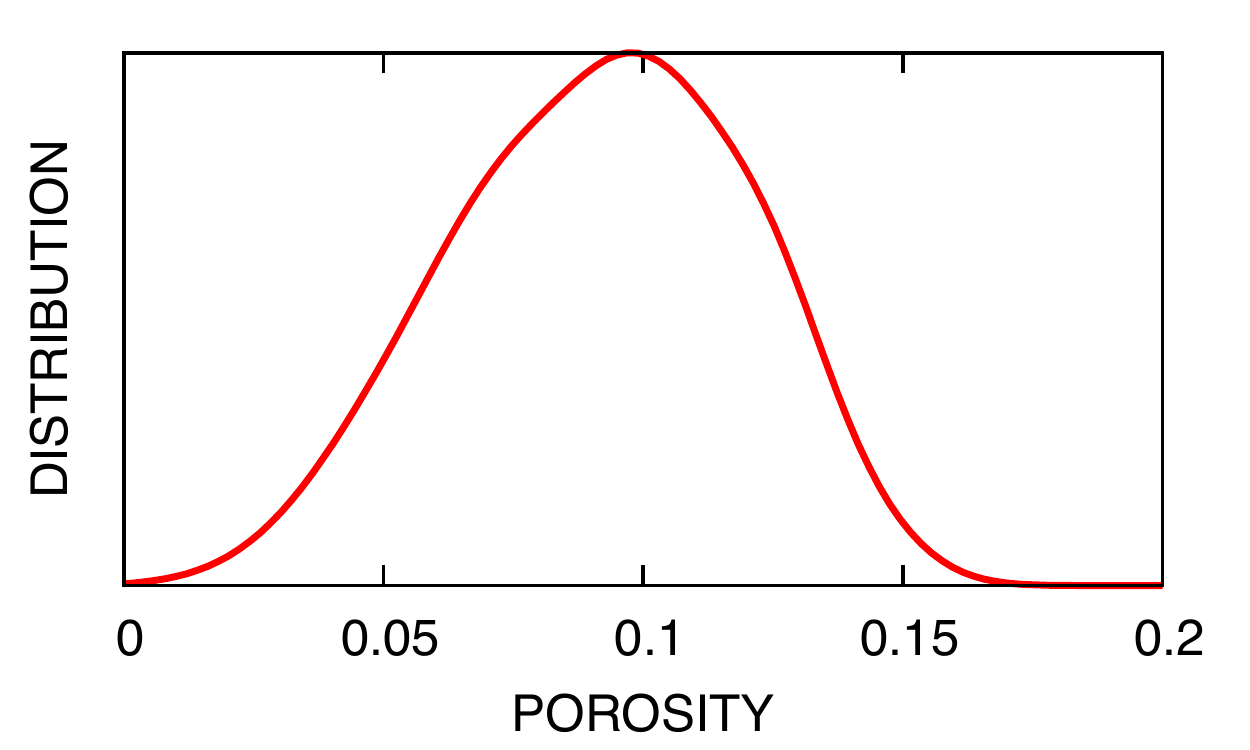}}
\caption{\ADD{Porosity distribution.}
}
\label{fig:porosity_distribution}
\end{figure}

\ADD{
\fref{fig:porosity_distribution} shows the distribution of sample porosities over an ensemble of 1121 realizations.
The distribution follows the target beta distribution with 0.09 mean porosity and a standard deviation of 0.03.
}

\begin{figure}
\centering
{\includegraphics[width=0.95\textwidth]{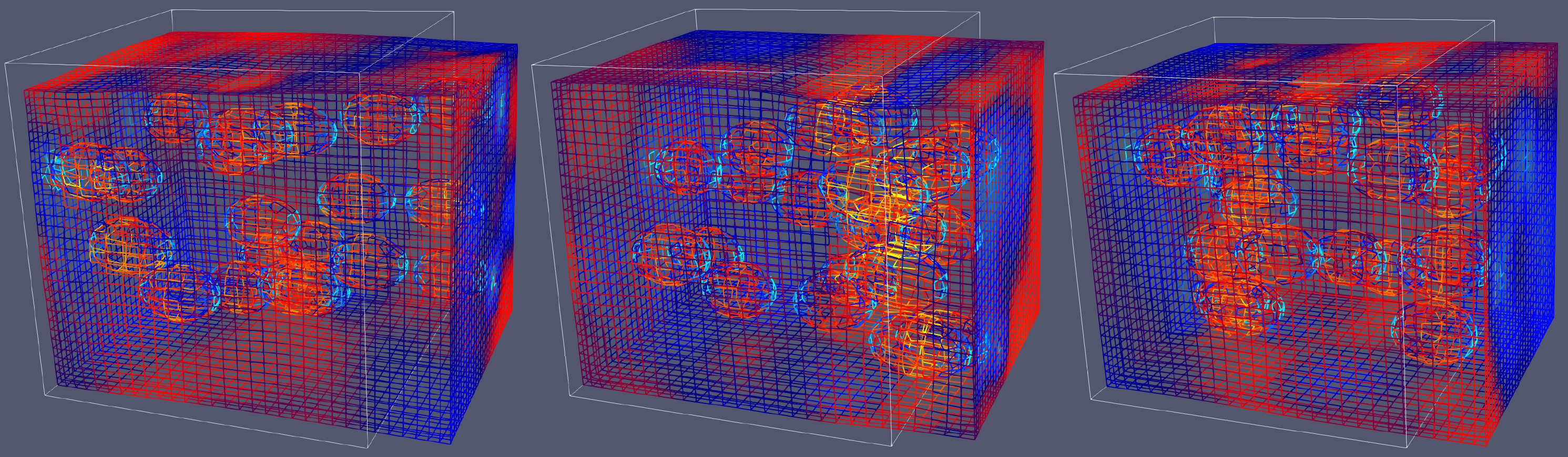}}
\caption{\ADD{Pore realizations showing exterior and interior surface mesh at 20 \% strain colored by tensile stress (blue: $<$ 0, red: 700 MPa).
The original, undeformed configuration is outlined.}
}
\label{fig:pore_stress}
\end{figure}

\begin{figure}
\centering
{\includegraphics[width=0.55\textwidth]{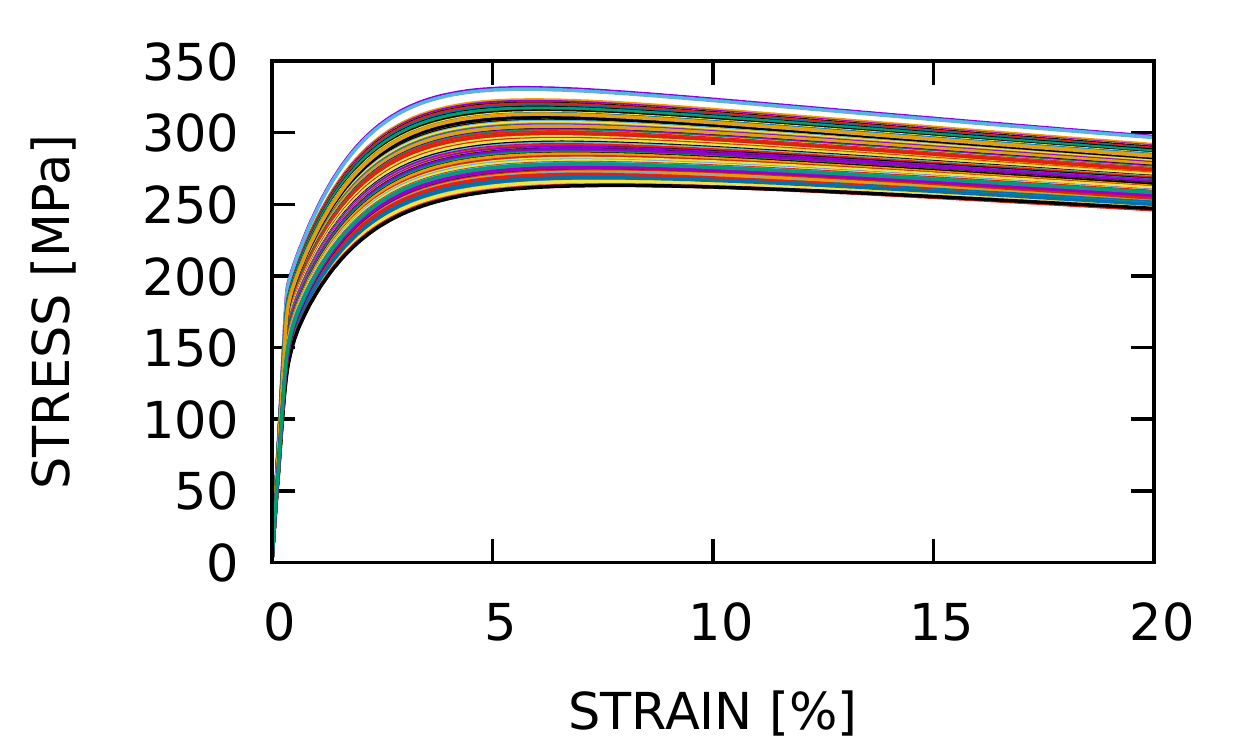}}
\caption{\ADD{Porous elastic-plastic stress response.
Color distinguishes the 64 realizations shown.
}
}
\label{fig:porous_stress_history}
\end{figure}

\ADD{
Due to variation in the number of voids, their size, and their placement relative to each other and inside the geometry, the response varies as the pores decrease the material stiffness and neighboring pores increase local stress concentrations.
The three representative realizations shown in \fref{fig:pore_stress} at the final strain of 20\% display significant heterogeneities in their deformation due to the dense packing of large pores.
The middle realization, in particular, clearly shows a plastic localization plane  due to a collection of pores leading to a weak section in the sample.
The stress fields for each of the samples display correspondingly large local variations and gradients.
The average stress histories, $\{ \sigma(t) \}$, shown in \fref{fig:porous_stress_history} display a variation of 22\%.
After rescaling by a mixture rule based on the solid fraction, 10\% variation remains.
This indicates that the details of the pore configurations control a significant portion of the plastic response.
}

\subsubsection{Crystal plasticity} \label{sec:cp_response}
\begin{figure}[htb!]
\centering
\includegraphics[width=0.9\textwidth]{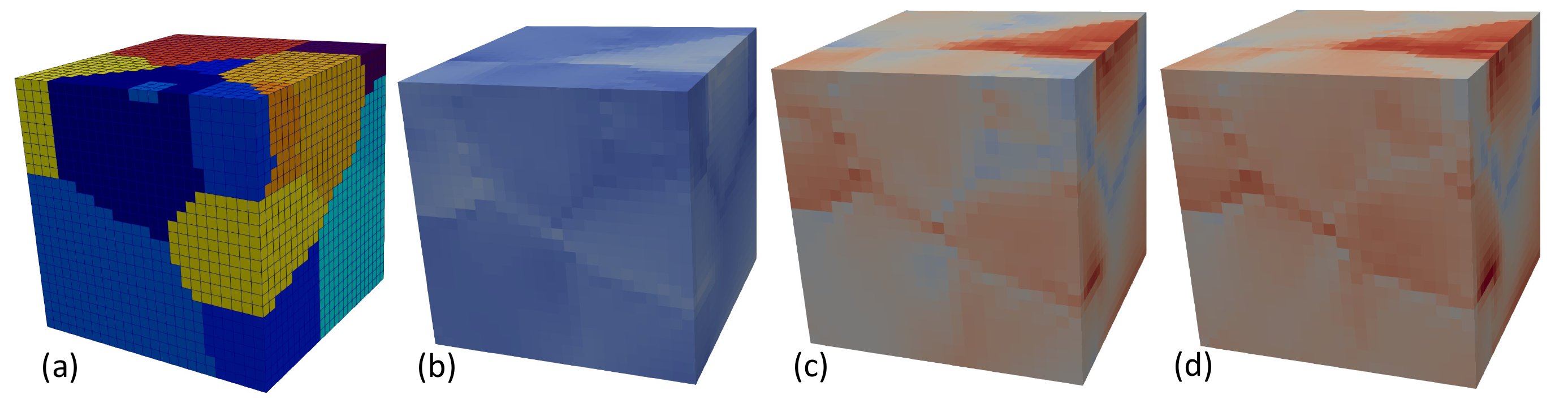}
\caption{\ADD{Polycrystal orientation (a) colored by the first Euler angle, and stress states: (b) elastic, (c) transition, (d) plastic, colored by $\sigma_{11}$ with the same scale for all three panels (blue: 0 MPa, red: 250 MPa).
}}
\label{fig:stress_field}
\end{figure}

The CP microstructure also evokes a complex and evolving stress field, as \fref{fig:stress_field}b--d illustrates.

\begin{figure}
\centering
\includegraphics[width=0.5\textwidth]{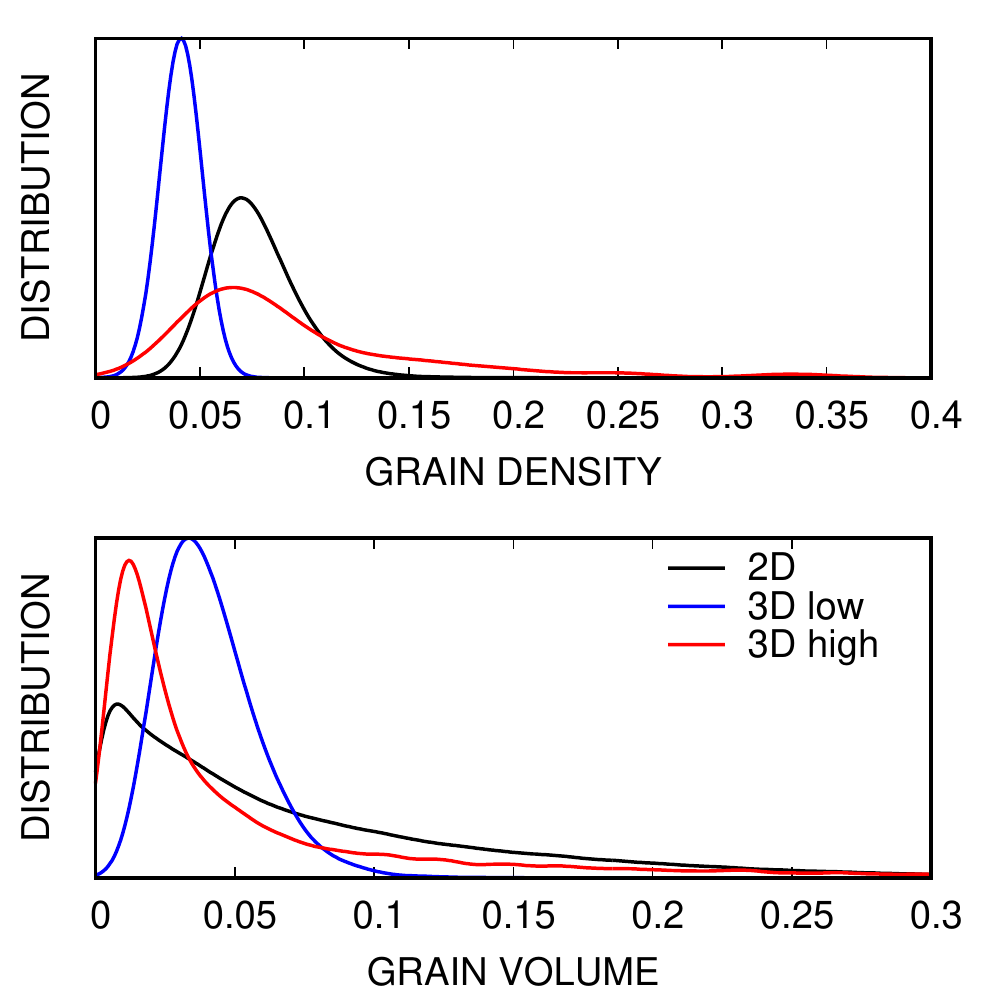}
\caption{Comparative statistics for 2D and 3D high and low variance CP ensembles.
Grain densities per realization and volumes per grain have been referenced to unit cell volumes in 2D and 3D, respectively.
}
\label{fig:ensemble_statistics}
\end{figure}

\begin{figure}
\centering
\includegraphics[width=0.5\textwidth]{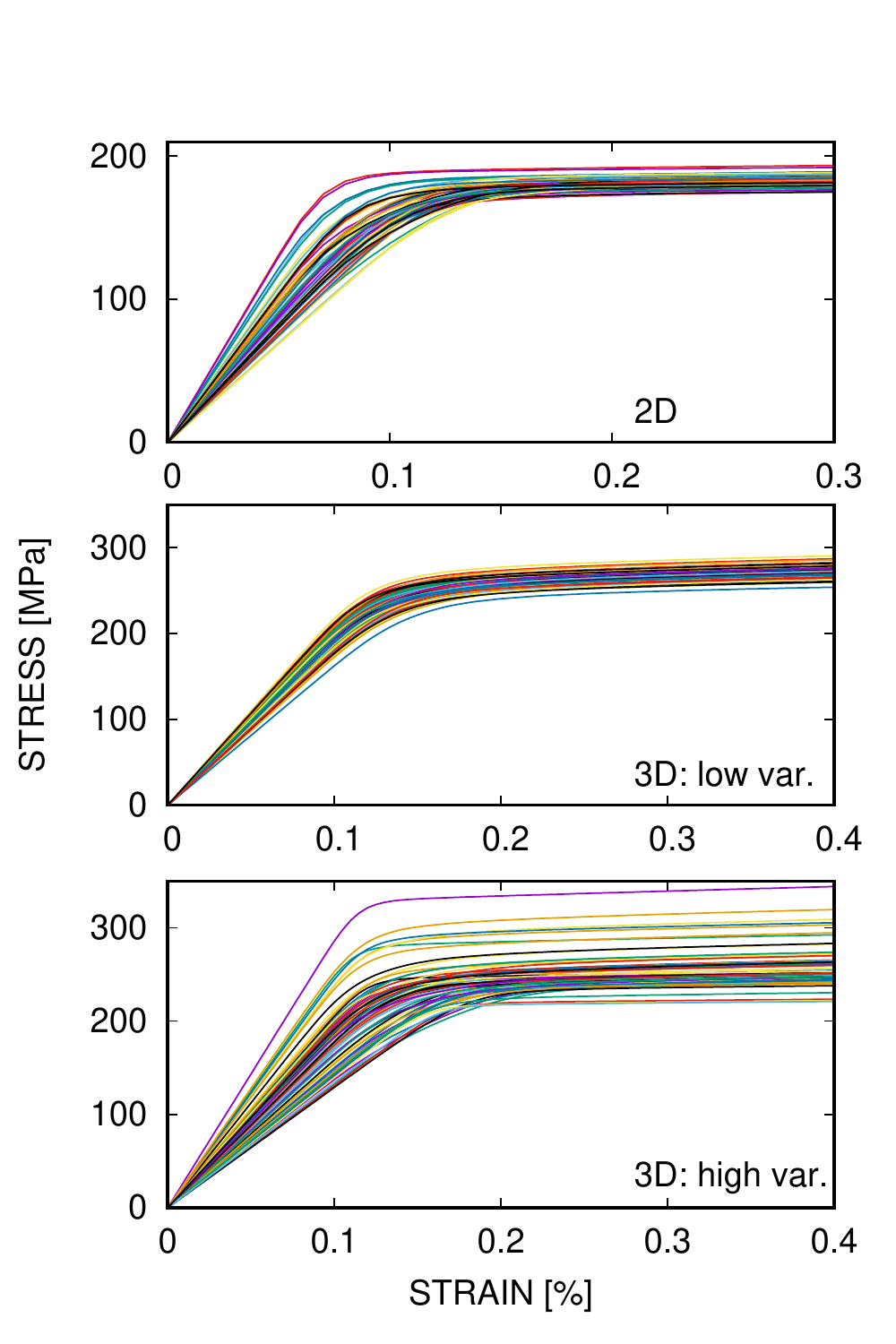}
\caption{CP stress response for 2D (top), 3D low grain size variance (middle), and 3D high variance (bottom) ensembles.
Color distinguishes the 64 realizations shown.
}
\label{fig:stress_history}
\end{figure}

Descriptive statistics of the 2D and 3D ensembles are shown in \fref{fig:ensemble_statistics}.
The distribution of grain densities of each realization (the reciprocal of the number of grains in the particular realization) is relatively broad and long-tailed in the high variance 3D dataset compared to that of the low variance 3D dataset.
The grain distribution of the 2D dataset has more compact support relative to the high variance 3D ensemble but is significantly wider than the low variance 3D ensemble, thus representing an intermediate distribution.
The distribution of the individual grain volumes over all realizations show similar trends.
The peak width of the high variance 3D dataset, however, is relatively narrow compared to the high variance 3D dataset.
The 2D data has a pronounced tail and indistinct peak which indicates a wide variance in grain sizes across the ensemble.
\fref{fig:stress_history} illustrates how the variance of the inputs $\phib(\Xb)$ is reflected in the variance of the output $\sigma(t)$.

These datasets are particularly challenging to represent, relative to existing work, since each realization had a distinct topology/grain assignment and texture.
Previous work \cite{frankel2019oligocrystals,vlassis2020geometric} employed a limited number of grain topologies \ADD{(the tiling of the domain by distinct subregions)} with unique texture assignments \ADD{(the specific orientations assigned to the grains)}.
Specifically, here each dataset had on the order of 10,000 unique topologies, whereas  in \crefs{frankel2019oligocrystals,vlassis2020geometric} there were on the order of 10--100 topologies.
\ADD{
It is plausible, in a small dataset, when the number of convolutional filters approaches the number of microstructural topologies a simpler learning process results since each filter can specialize to a particular topology.
In this study this is not the case since each of the 10$^4$ samples had both a unique grain structure and grain orientation (texture).
Clearly, with datasets of this size and variety, the filters must learn generalized, predictive features.
}

\section{Neural network architecture} \label{sec:architecture}

The overall neural network (NN) architecture to model the problem of interest \eref{eq:problem} is analogous to the hybrid network from our previous work \cite{frankel2019oligocrystals,frankel2019evolution} and also used in \cref{vlassis2020geometric}.
It is illustrated in \fref{fig:architecture} and consists of two main components: (a) a convolutional neural network (CNN or GCNN, orange) to process the spatially complex initial state $\phib(\Xb)$ and (b) a recurrent neural network (RNN, blue) to evolve the quantity of interest $\sigma(t)$ given the time-dependent loading $\epsilon(t)$ as an input.
As can be seen in \fref{fig:architecture}, the output features of the CNN become inputs to the RNN along with the time-dependent loading.
This hybrid NN has the potential to be quite complex with many hyper--parameters, \eg a different kernel width for each convolution layer and different number of nodes for each dense layer.

In this work we simplify the NN architecture from that we previously employed to facilitate exploration and comparison of CNN and GCNN approaches.
This is not particularly constraining since there are redundancies in the approximation power of such a network.
We determine the shape of the network diagrammed in \fref{fig:architecture} with three hyper-parameters:
(a) the number of filters, $\Nf$, applied in parallel to node data;
(b) the number of convolutional layers, $\Nc$;
and
(c) the number of densely connected layers, $\Nd$, post-convolution used in the reduction of image to (hidden) features.
In the following we will use the abbreviation $\arch{\Nf}{\Nc}{\Nd}$ to refer to architectures defined by these three parameters, for instance $\arch{4}{2}{1}$ is a network with 4 filters, 2 convolutional layers, and 1 dense layer.
Note the global (average) pooling after the convolutional layers reduces the output of the image processing CNN to $\Nf$ outputs and hence determines with width of the entire convolutional component including the densely connected layers.
Note, in our previous work \cite{frankel2019oligocrystals} we used an encoder for this task (c).
\ADD{
\fref{fig:cnn} illustrates the details of a typical convolutional component in the overall architecture (yellow in \fref{fig:architecture}).
}
A batch normalization layer (not shown) is inserted between each convolutional layer to aid in conditioning the output and training \cite{bjorck2018understanding}.
To facilitate comparison of the CNN and GCNN approaches we fix the kernel width of pixel convolutional layer to 3 to make an analog of a graph where connections are nearest neighbors for the physical problem.
The particular RNN we employ is the well-known long-short-term memory unit (LSTM) \cite{hochreiter1997long}.
In preliminary studies we also tried another standard RNN, the gated recurrent unit (GRU) \cite{cho2014learning}, which achieved marginally worse accuracy on average.
The RNN used $\tanh$ activations and all other layers employ rectifying linear units (ReLUs) for their nonlinear activation functions.
We also tried $\tanh$ activations for the entire network but that configuration choice performed relatively poorly \cite{glorot2010understanding,glorot2011deep}.
Lastly we apply the standard technique of using a linear mixing layer (no nonlinearity, only affine transformation) just prior to output of interest.

The primary variation in the networks we explore is in the convolutional unit (yellow in the schematic \fref{fig:architecture} and shown in detail in \fref{fig:cnn}) and in particular the construction of the convolution filter, which is our focus.

\subsection{Proposed graph structure}
At the level of digitized data in a pixel-based CNN, the microstructural input is values of $\phib$ at pixels (or a cell or an element) that are addressed in a grid-wise fashion.
In general the field $\phib(\Xb)$ can be represented as a pixelized image on a structured grid or on an unstructured mesh.
For a graph representation, such as that used in Vlassis \etal \cite{vlassis2020geometric}, the reduction of a clearly segmentable microstructure, such as that illustrated in \fref{fig:stress_field}a, leads to nodes representing homogeneous regions and graph edges encoding adjacency between regions.
This reduction loses spatial information such as the shape of the regions in the clustering/aggregation to nodes.
Hence, that framework typically requires enrichment of the node data $\xs$ by featurization, \ie picking measures/statistics that quantify the information lost in the reduction.
In that approach, the number of nodes $N_\text{nodes}$ is variable, and equal to the number of regions in the particular sample.
In that context, and in general, the node data $\xs$ is number of nodes $N_\text{nodes}$ by number of input features  $\Nf$ derived from $\phib(\Xb)$.
Some features are obvious, such as the value of $\phib$ in the clustered region represented by the node and the volume of the region, and others are not.
It is easy to see that these features can result from preconceived filters and clustering operations applied to image $\phib(\Xb)$.
We will refer to this feature-based, reduced graph convolutional network as a \RGCNN.

To avoid nebulous task of featurization, we propose a direct graph CNN (\DGCNN) where $\xs$ is identically $\phib(\Xb)$ at the cell/element centers of the computational grid but flattened (in an arbitrary order) to fit in the graph convolution paradigm.
The graph convolutions are permutationally invariant so ordering does not affect output.
In the proposed network $N_\text{nodes}$ is the number of pixels or unstructured elements in the image and the edges are derived directly from the mesh topology, \ie element/pixel neighbors are graph neighbors.
This approach has qualitative advantage of being a graph-based representation, which has intrinsic invariance properties, while working directly on the structured or unstructured data $\phib(\Xb)$ not a segmented or clustered version of it.
This representation does not preclude the use of derived, informative features to boost accuracy by adding them to $\xs$, as we will demonstrate.

The proposed architecture was implemented with Spektral \cite{spektral} and TensorFlow \cite{tensorflow}.

\subsection{Comparison of pixel and graph based convolution}
\begin{figure}
\centering
\includegraphics[width=0.7\textwidth]{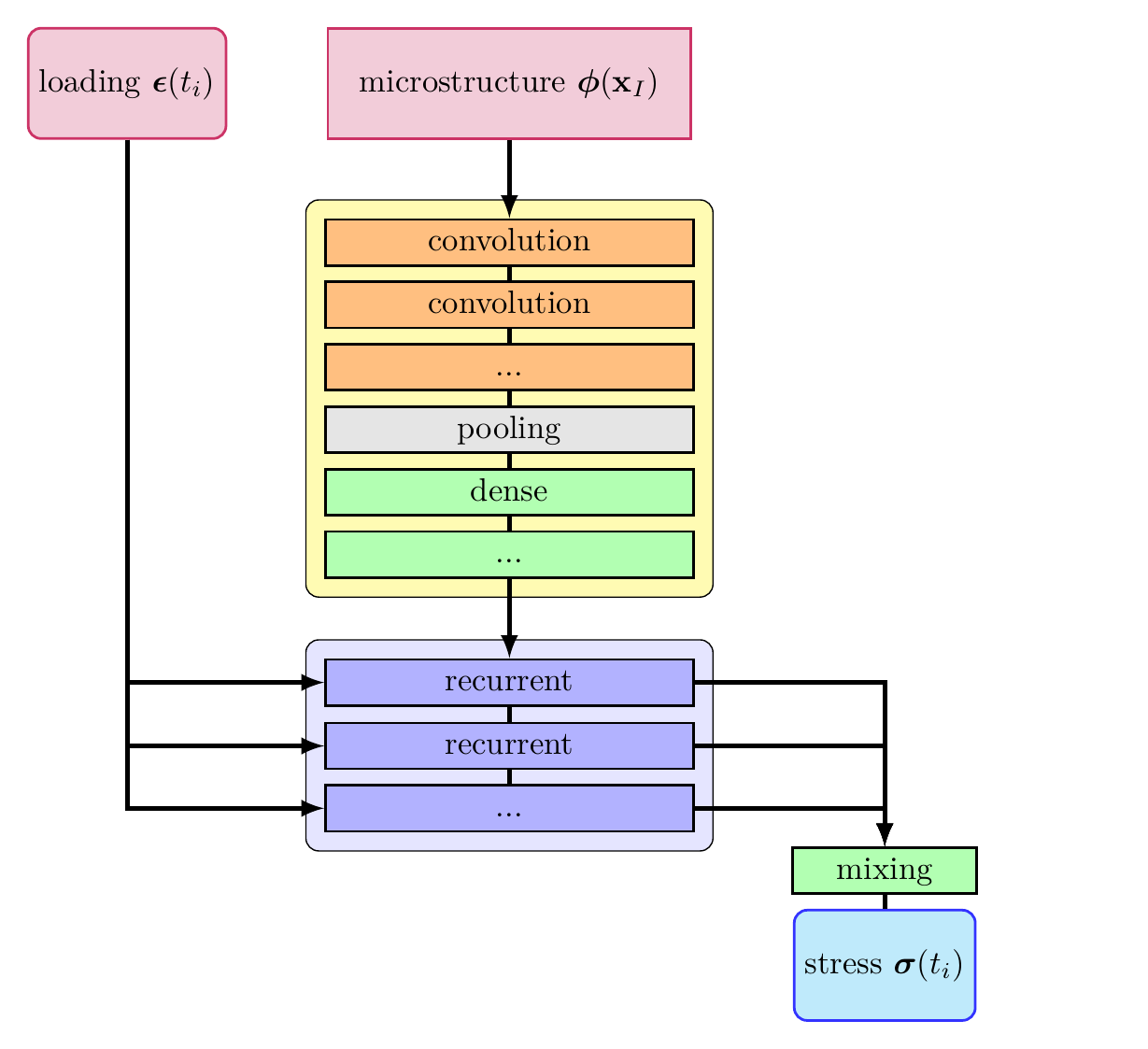}
\caption{Hybrid neural network with CNN (yellow) and RNN (blue) components to transform the inputs (red): spatially-dependent $\phib(\Xb)$ microstructure and time-dependent loading history $\epsilonb(t)$, to the output $\sigmab(\phib,t)$ (cyan).
}
\label{fig:architecture}
\end{figure}

\begin{figure}
\centering
\includegraphics[width=0.85\textwidth]{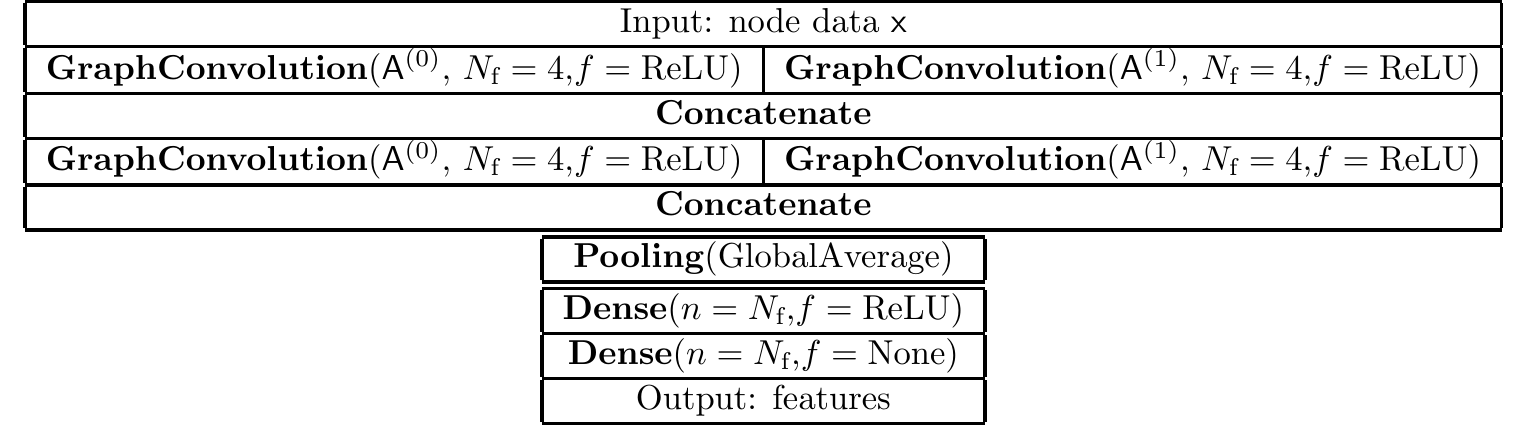}
\caption{\ADD{A \arch{4}{2}{2} convolutional unit.
$\As^{(i)}$ is the adjacency matrix for the $i$-th neighbors (0 is self), $N_\text{f}$ is the number of filters, and $f$ is the activation function for the layer.
Note the last dense layer is linear (no non-linear activation function).
} }
\label{fig:cnn}
\end{figure}

\begin{figure}
\centering
\begin{subfigure}[c]{0.3\textwidth}
\includegraphics[width=1.0\textwidth]{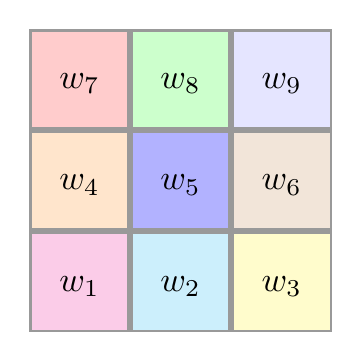}
\caption{pixel}
\end{subfigure}
\begin{subfigure}[c]{0.3\textwidth}
\includegraphics[width=1.0\textwidth]{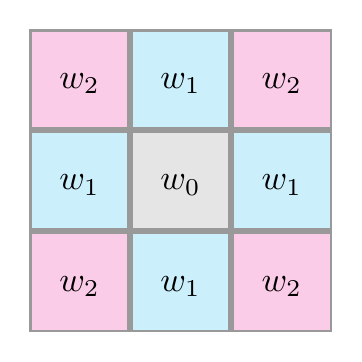}
\caption{$\ast$}
\end{subfigure}
\begin{subfigure}[c]{0.3\textwidth}
\includegraphics[width=1.0\textwidth]{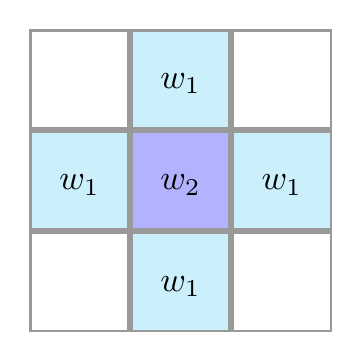}
\caption{\#}
\end{subfigure}
\begin{subfigure}[c]{0.3\textwidth}
\includegraphics[width=1.0\textwidth]{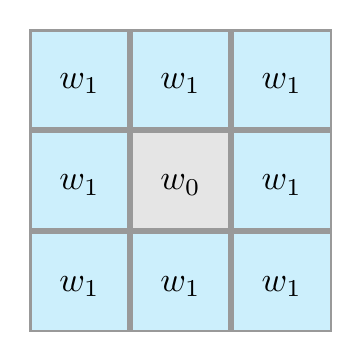}
\caption{O}
\end{subfigure}
\begin{subfigure}[c]{0.3\textwidth}
\includegraphics[width=1.0\textwidth]{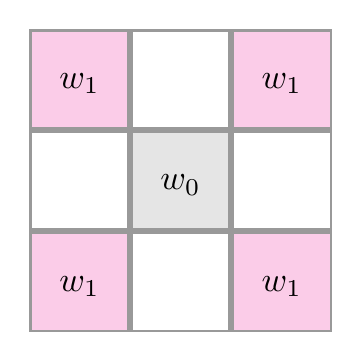}
\caption{X}
\end{subfigure}
\begin{subfigure}[c]{0.3\textwidth}
\includegraphics[width=1.0\textwidth]{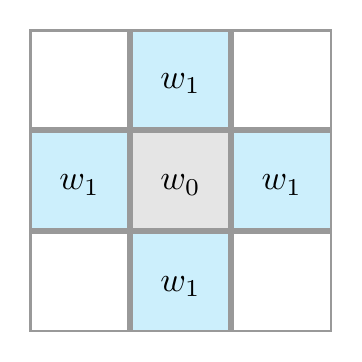}
\caption{+}
\end{subfigure}
\caption{Convolutional filters in 2D:
(a) CNN where all neighbors have independent weights (denoted by colors); and
alternative GCN-like filters that preserve invariant outputs:
(b) $\ast$: edges and nodes have separate weights
(c) \#: all edges have a single weight.
(d) O: all neighbors have a single weight.
(e) X: all node neighbors have single weight.
(f) +: all edges have a single weight.
Edge neighbors share an edge/face, whereas node neighbors only share a vertex/node.
Color: trainable, gray: determined by trainable, white: not used.
}
\label{fig:filters}
\end{figure}

To understand the difference between a pixel-based and a graph-based convolution, let us examine a single convolutional filter.
Note that the proposed architecture shown in \fref{fig:architecture} and \fref{fig:cnn} has several filters.
In general, a convolutional filter has trainable weights $\Ws$ and bias $\bs$.
The weight matrix $\Ws$ effects an affine transform by matrix multiplication that mixes input features $\xs$ into output features $\ys$ and $\bs$ provides an offset to tune activation of the subsequently applied nonlinearity $\ys=f(\xs)$.
In a traditional grid/pixel convolution (CNN), the node features are addressable by grid index, for instance in 2D:
\begin{equation} \label{eq:conv2d}
\ys_{(i,j)} = \sum_{k',l'} \Ws_{(k',l')} \xs_{(i+k',j+l')} + b \ ,
\end{equation}
where $i,j$ are indices over the pixelated image/discretized field, $k',l'$ indices over the convolution/filter which is $N_\text{kernel} \times N_\text{kernel}$ in size, with $N_\text{kernel} \ll N_\text{nodes}$ being the kernel width.
To see the similarities with graph convolution we recast this convolutional multiplication in \eref{eq:conv2d} by mapping both $(i,j)$ and  $(k',l')$ to single indices $I$ and $L$ by imposing a fixed, arbitrary ordering across the kernel $\Ws$ (refer to \fref{fig:filters}).
For the flattened input $\xs$, the corresponding output $\ys$ is
\begin{equation} \label{eq:gconv}
\ys_I = \sum_J \left[ \sum_{K'} \Ws_{K'}  \As_{IJ}^{(K')} \right] \xs_J + \bs_{K'} \ ,
\end{equation}
where multiplication by a matrix $\As^{(K')}$ provides a masking operation translating the global indices to local dependencies.
Here $\As^{(K')}$ is a global $N_\text{nodes} \times N_\text{nodes}$ adjacency matrix for each entry $K'$ in the convolutional kernel (\ie each pixel under the filter kernel is treated uniquely);
and $\As_{IJ}^{(K')} = 1$ if $I$ and $J$ are neighbors (by some definition \eg shared face, shared node in the computational mesh or distance) and 0 otherwise.
The definition of neighbors determines the direct interactions, in rough analogy to choosing the kernel width in a pixel-based convolutional filter.

Likewise, for a graph convolutional network (GCN) layer \cite{kipf2016semi}, the convolution operation takes the form:
\begin{equation}
\ys_I = \sum_J W \As_{IJ} \xs_J + b \ ,
\end{equation}
where the adjacency $\As$ plays the same masking/connectivity role as $\As^{(K')}$ and an ordering of the input data $\xs$ is necessary to associate the indices $I$ and $J$ with particular cells.
Again $\As_{IJ} = 1$ if $I$ and $J$ are neighbors by some definition and 0 otherwise.
Based on the derivation of the GCN, refer to \aref{app:gcn}, self-loops are added such that $\As_{II} = 1$.
For each filter there are two trainable parameters, $W$ and $b$, versus $k^d$ for a pixel-based CNN filter (where $d$ is the spatial dimension and $k$ is the kernel width).
These graph-based filters have permutational invariance by construction since all neighbors have same weight.
They are also typically normalized by the number of neighbors, for instance in a GCN, the binary adjacency matrix $\tilde{\As}$
is normalized by the degree matrix $\Ds_{IJ} = \sum_I \tilde{\As}_{IJ} \delta_{IJ}$
\begin{equation}
{\As} = \Ds^{-1/2} \tilde{\As} \Ds^{-1/2}
\end{equation}
to convey average neighborhood information and improve the conditioning of $\As$.

Finally, in both the pixel and graph convolutional cases an activation function $f$ is applied element-wise to the resulting node data $\ys$
\begin{equation}
\xs^\dag = f\left(\ys(\xs)\right)
\end{equation}
to obtain the input $\xs^\dag$ to the next layer from that of the present layer $\xs$.

Clearly the adjacency $\As$ plays a crucial role in convolution by defining what constitutes influence between nodes.
In physical problems typically influence is local and decays with distance.
For some applications, such as electrostatic interactions the interactions do not decay quickly with distance and hence are long-ranged \cite{li2020multipole}.
Most other applications have relatively short-ranged influence, such as contact/interface interactions.
As mentioned, the source field for the microstructure is pixelated image grid or computational mesh which have their own obvious topology and neighbors.
In the proposed direct graph we define adjacency by the pixel/cell neighbors of the image grid/mesh not by the data $\phib(\Xb)$ on it.
When reduced to a graph in this fashion, where nodes are the pixels/cells, the only topologically distinct nearest neighbors are: (a) those that share an edge with the particular cell or (b) only a vertex.
In CNNs compact kernels are preferable since they limit the number of parameters that need to be trained.
For GCNNs, and in particular the GCN, neighbors are not distinct and hence there is only a single weight.
As in CNNs longer range influence can be captured by multiple layers applied in sequence.

The Kipf and Welling~\cite{kipf2016semi} GCN is well-known to be an effective (non-spectral) convolutional filter.
By comparing it to standard CNNs we employed in our previous work~\cite{frankel2019predicting} we devised a few variants based on manipulating the node adjacencies and associated weights.
\fref{fig:filters} shows the variety of rotationally and permutationally invariant filters we explored.
Note these filters assume that the neighbors are at equivalent distances from the node at the center and the cells are comparable sizes.
\ADD{
We generalize this type of filter to multiple adjacencies $ \As^{(K')}$ with their associated weights $\Ws_{K'}$ via:
\begin{equation} \label{eq:gcnn}
\ys_I = \sum_J \left[ \sum_{K'} \Ws_{K'}  \As_{IJ}^{(K')} \right] \xs_J  \ .
\end{equation}
}

The pixel-based filter in a standard CNN treats every pixel in the kernel independently (refer to \fref{fig:filters}a) and retains a sense of how pixels located relative to the central pixel.
This richness has the side effect of not satisfying the symmetries required by invariance.
\ADD{
To elaborate with an illustration, if the data $\{ \phib, \epsilonb; \sigmab \}$ is rotated by $\pi/2$ around a coordinate axis the data at pixels are uniquely mapped by $\Qs$ to new indices without interpolation $\xs^* = \Qs \xs$.
In this simple case the linear transformation
\begin{equation}
\Qs \ys_I \equiv
\Qs \sum_J \left[ \sum_{K'} \Ws_{K'}  \As_{IJ}^{(K')} \right] \xs_J
\neq
\sum_J \left[ \sum_{K'} \Ws_{K'}  \As_{IJ}^{(K')} \right]^* \Qs \xs_J
\end{equation}
in \eref{eq:gconv} applied to the inputs $\xs_J =  \phi(\Xb_J)$ will not produce a rotated output since image $\xs = \phi(\Xb)$ rotates  but the filter $\left[ \Ws_{K'} \As^{(K')}\right]$ for the $K'$-th adjacency does not.
A graph treatment, unlike a pixel convolution, effectively allows the transformation $\Qs$ to commute with the adjacency, $ \Qs \As^{(K')} \xs = \left[\As^{(K')} \right]^* \Qs \xs$, since the node association of the graph adjacency is invariant to spatial transformations.
Refer to \crefs{kondor2018generalization,finzi2020generalizing} for a more formal treatment.
The equivariance problem remains for the action of the weights $\Ws_{K'}$.
}

Permutational invariance of the weights is one means of solving this issue.
Strictly speaking, the  weights of all of the  neighbors of the central pixel are required to be identical for permutational invariance (and for invariance with respect to arbitrary rotations).
We implemented filter variants where edge and vertex weights are used exclusively \ADD{(\fref{fig:filters}f uses edge neighbors whereas \fref{fig:filters}e employs only node neighbors)} or given independent weights, such as the ``$\ast$'' pattern in \fref{fig:filters}b.
\ADD{
These filters are in contrast to the CNN filter, shown in \fref{fig:filters}a, which has no inherent symmetries.
}
Note there is some evidence that with sufficient data CNNs learn rotational invariance \cite{quiroga2018revisiting} but the benefits of a smaller parameter space and a compact representation that satisfies this constraint exactly are clear.
In the GCN, as designed by Kipf and Welling \cite{kipf2016semi}, the weight of the center (shown in gray in \fref{fig:filters}f \vs colored in \fref{fig:filters}c) was chosen to be the same as the neighbors (refer to \aref{app:gcn}).
In this work we also tried variants where the self-weight is independent of the neighbor weight (\fref{fig:filters}c \vs \fref{fig:filters}f).
\ADD{
As in \eref{eq:gconv}, each independent, trainable weight $\Ws_{K'}$ can be associated with a different pre-determined, binary adjacency matrix $\As^{(K')}$, or, equivalently, the weighted adjacency  $\sum_{K'} \As^{(K')}_{IJ} \Ws_{K'}$ can be considered the trainable entity.
}

\section{Results}\label{sec:results}

\ADD{
To establish the efficacy of the proposed GCNN-RNN architecture  applied directly to unstructured mesh data without featurization  we demonstrate its performance on the porous material dataset.
}
Since hyper-parameter optimization and other computationally intensive tasks are more feasible with the 2D dataset, we use it explore a variety of hyper-parameters and architecture choices with the 2D dataset.
After determining what graph-based filters perform well and have good accuracy per parameter and dataset size, we demonstrate the proposed architecture on the two 3D CP datasets.

The data was conditioned to aid training.
The input data $\phib(\Xb)$ and $\epsilon(t)$ were normalized by their maximum values since both had lower bounds of zero.
The output data $\sigma(\phib_a,t)$ was transformed to the difference $\Delta \sigma(\phib_a,t)$ between the data $\sigma(\phib_a,t)$ and its mean trend $\langle \sigma \rangle = \frac{1}{N} \sum_a {\sigma}(\phib_a,t)$ ($N$ is the number of realizations) and normalized by the standard deviation of $\sigma(\phib_a,t)$  over time $t$.
We chose to train the model response $\hat{\sigma}(\phib_a,t)$ to the difference from the mean trend to emphasize the variation in response between microstructures $\phib_a$.
We evaluated the performance of each of the convolutional networks primarily with the root mean squared error (RMSE)
\begin{equation} \label{eq:error}
\error(t) = \frac{1}{\max \langle \sigma \rangle} \sqrt{
{ \sum_a (\hat{\sigma}(\phib_a,t) - \sigma(\phib_a,t) )^2 }
}
\end{equation}
relative to the maximum $\sigma$ over the dataset,
and the (Pearson) correlation coefficient
\begin{equation} \label{eq:correlation}
\corr(t)        = \frac{\sum_a \Delta \hat{\sigma}(\phib_a,t) \, \Delta \sigma(\phib_a,t) }
{ \sqrt{ \sum_a \Delta \hat{\sigma}(\phib_a,t)^2 \
\sum_a \Delta     {\sigma}(\phib_a,t)^2 }}
\end{equation}
of the normalized data.
In \eref{eq:error} and \eref{eq:correlation} the sum is over all realizations $a$ in the test set, and the NN model response is denoted as $\hat{\sigma}(\phib_a,t)$.
We used a 70/10/20 train/validation/test split for the smaller porous material dataset and a 80/10/10 split for the larger CP datasets.

Recall the abbreviated architecture naming convention $\arch{\Nf}{\Nc}{\Nd}$, which will be used throughout this section.

\subsection{Demonstration on unstructured mesh data} \label{sec:unstructured}

\ADD{
We used the smaller 3D porous material dataset to demonstrate that GCNNs applied directly to unstructured mesh data are effective at representing homogenized material behavior.
For this study we used a \arch{32}{4}{2} convolutional unit with ``O'' type filters that treat all nearest neighbor elements equally.
The field data $\phib(\Xb)$, in this case, is the binary density field (0: void, 1: metal) on the native unstructured computational mesh.
Since element and systems sizes varied across the ensemble, we augmented $\phib(\Xb)$ with the volumes.
The stress response $\bar{\sigma}(t)$ is comprised of 400 tensile loading steps to reach 20\% strain.
Since each realization has a different mesh and each adjacency matrix is large (on the order of 10$^5$-10$^6$ rows and columns)  albeit sparse, we trained the network by evaluating and updating the network weights one sample at a time, \ie a batch size of 1.
}

\begin{figure}
\centering
{\includegraphics[width=0.55\textwidth]{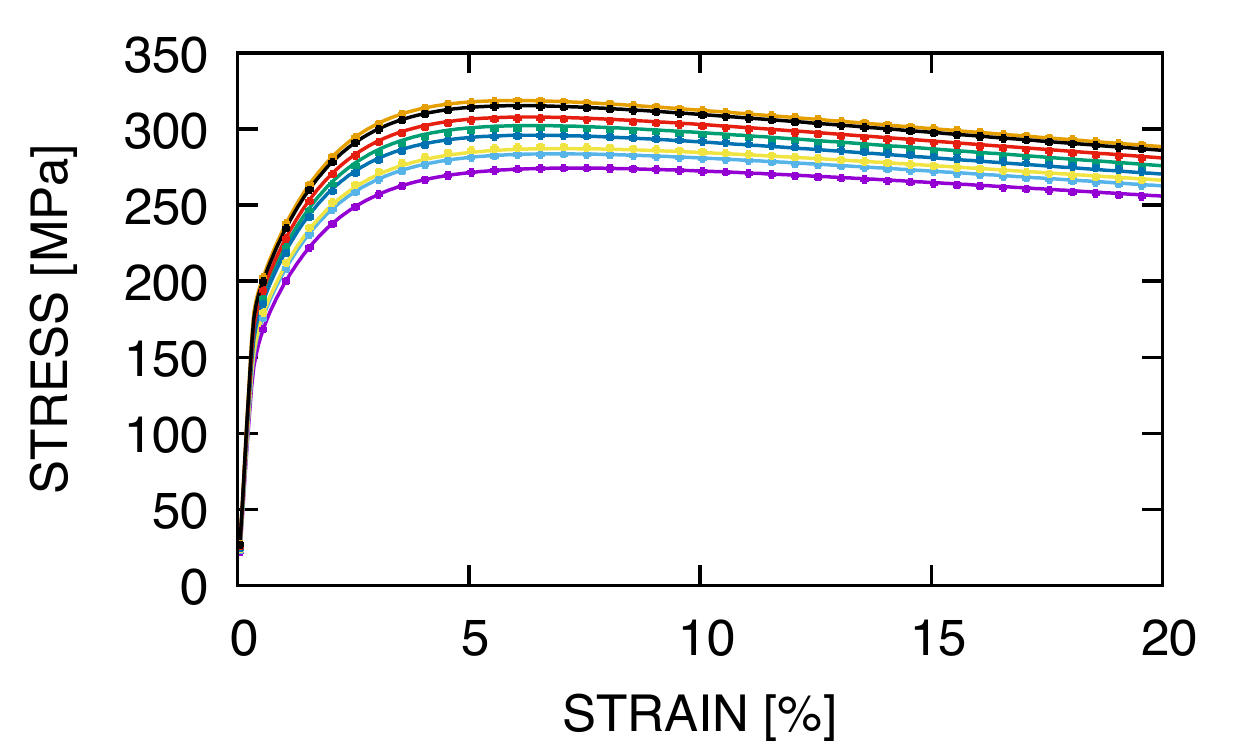}}
\caption{\ADD{Comparison of true (lines) and predictions (points with corresponding color) for 8 realizations of the porous metal data.}
}
\label{fig:pore_predictions}
\end{figure}

\ADD{
The predictions for 8 randomly selected trajectories shown in \fref{fig:pore_predictions} are smooth and display minimal errors that are fairly uniform across the evolution.
The initial elastic regime is well captured, as is the ultimate (peak) strength and subsequent plastic flow.
Over the entire test set of 224 samples, the normalized mean RMSE was 0.00409 with 0.00052 standard deviation, and the mean correlation was 0.995.
Clearly the architecture shown in \fref{fig:architecture} with graph convolution layers applied to unstructured field data is an effective model of this microstructural response.
}

\pagebreak

\subsection{Efficacy of the components of hybrid network} \label{sec:components}
In this study we vary the three chosen architecture hyper-parameters: $\Nf$, $\Nc$ and $\Nd$ for three architectures: (a) a CNN, (b) the proposed \DGCNN, and (c) a {\RGCNN} endowed with angle and volume features.
Both GCNN architectures employ the standard GCN filter, with a ``+'' pattern and dependent center weight illustrated in \fref{fig:filters}f.
\ADD{All models are trained to the 2D CP dataset.}

\begin{figure}
\centering
\begin{subfigure}[c]{0.9\textwidth}
{\includegraphics[width=0.45\textwidth]{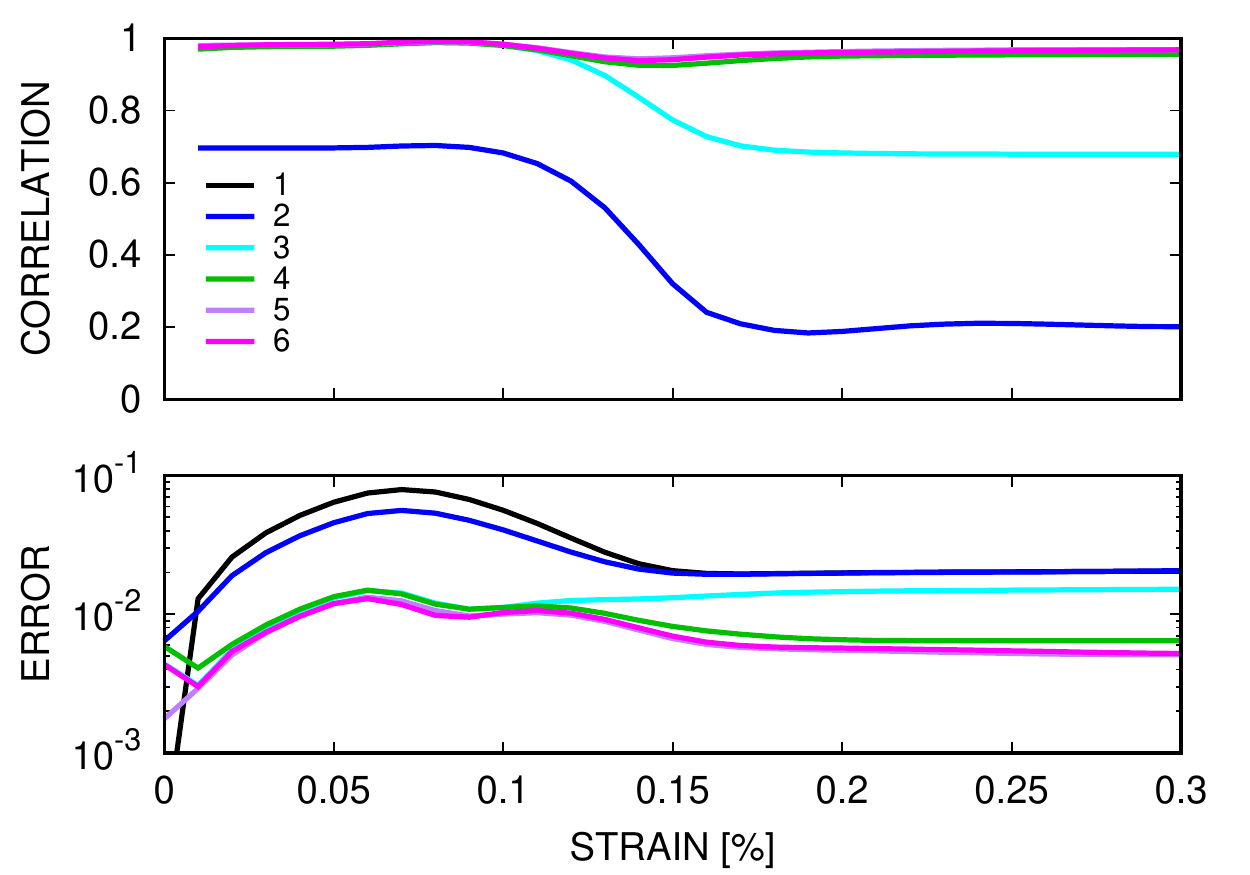}}
{\includegraphics[width=0.45\textwidth]{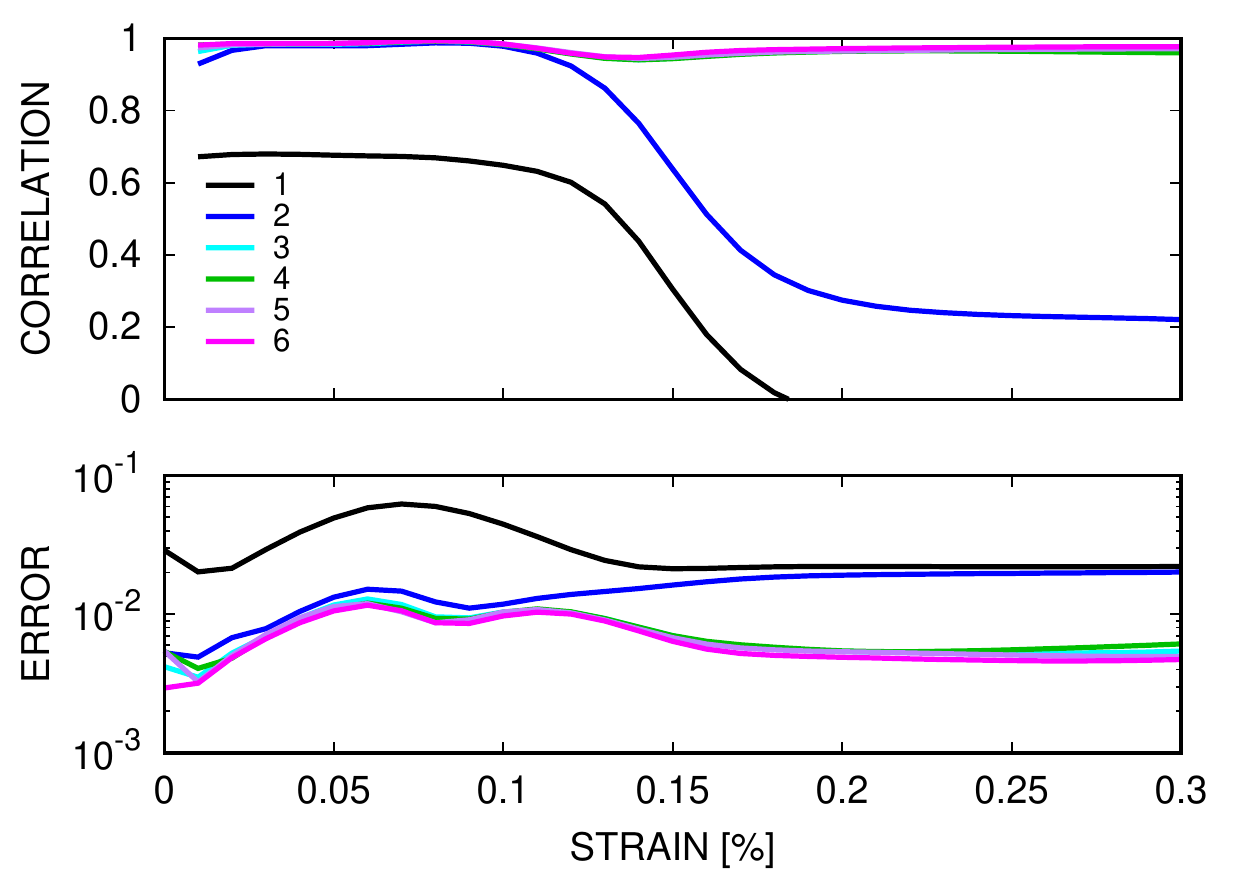}}
\caption{CNN}
\end{subfigure}
\begin{subfigure}[c]{0.9\textwidth}
{\includegraphics[width=0.45\textwidth]{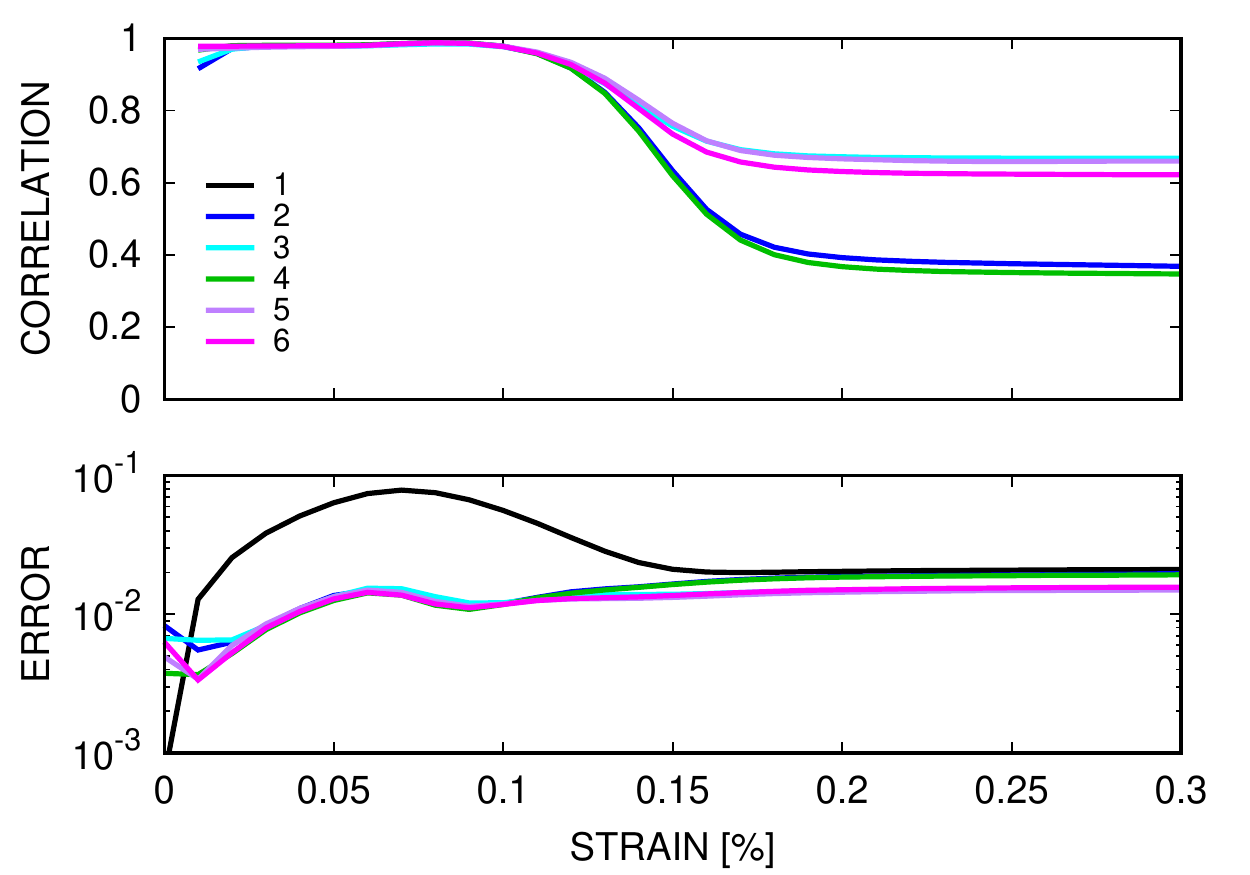}}
{\includegraphics[width=0.45\textwidth]{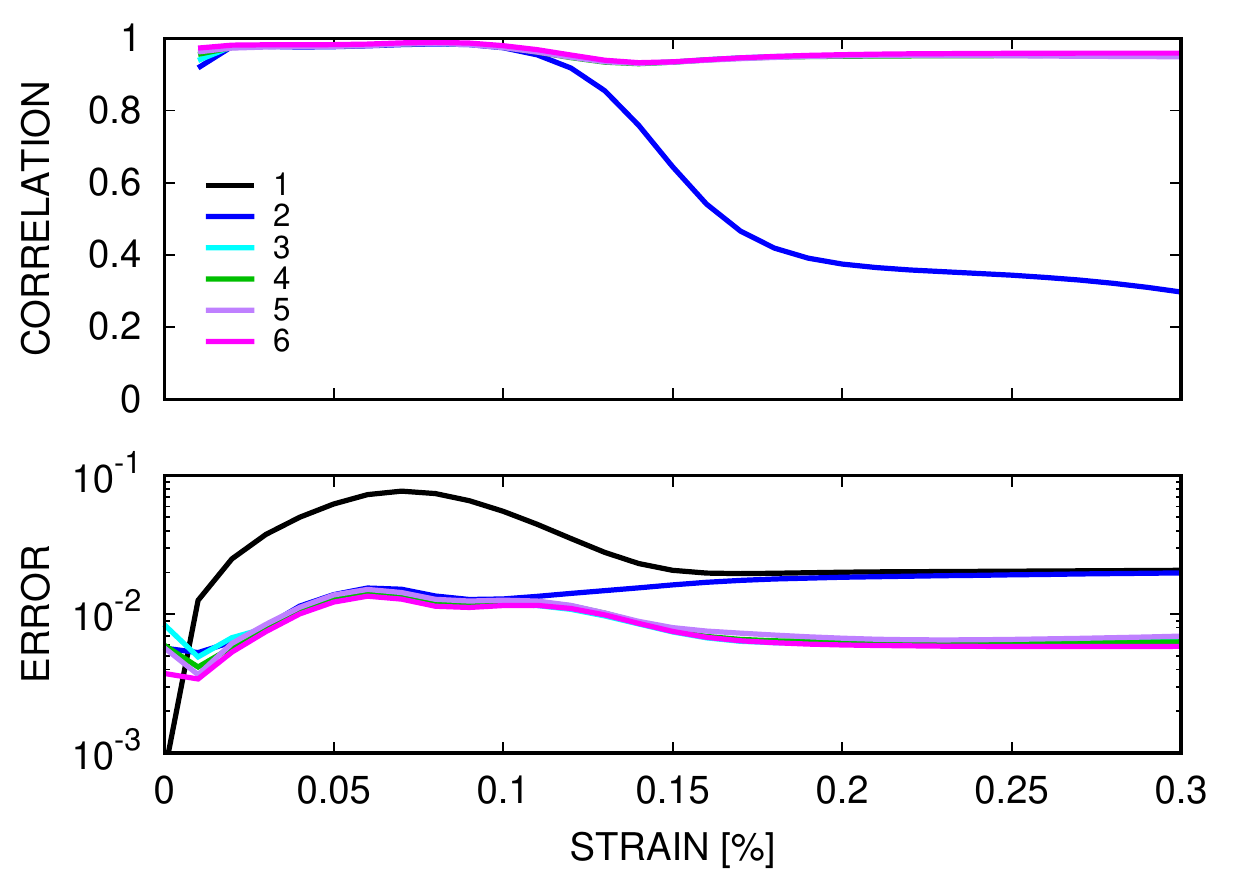}}
\caption{\DGCNN}
\end{subfigure}
\begin{subfigure}[c]{0.9\textwidth}
{\includegraphics[width=0.45\textwidth]{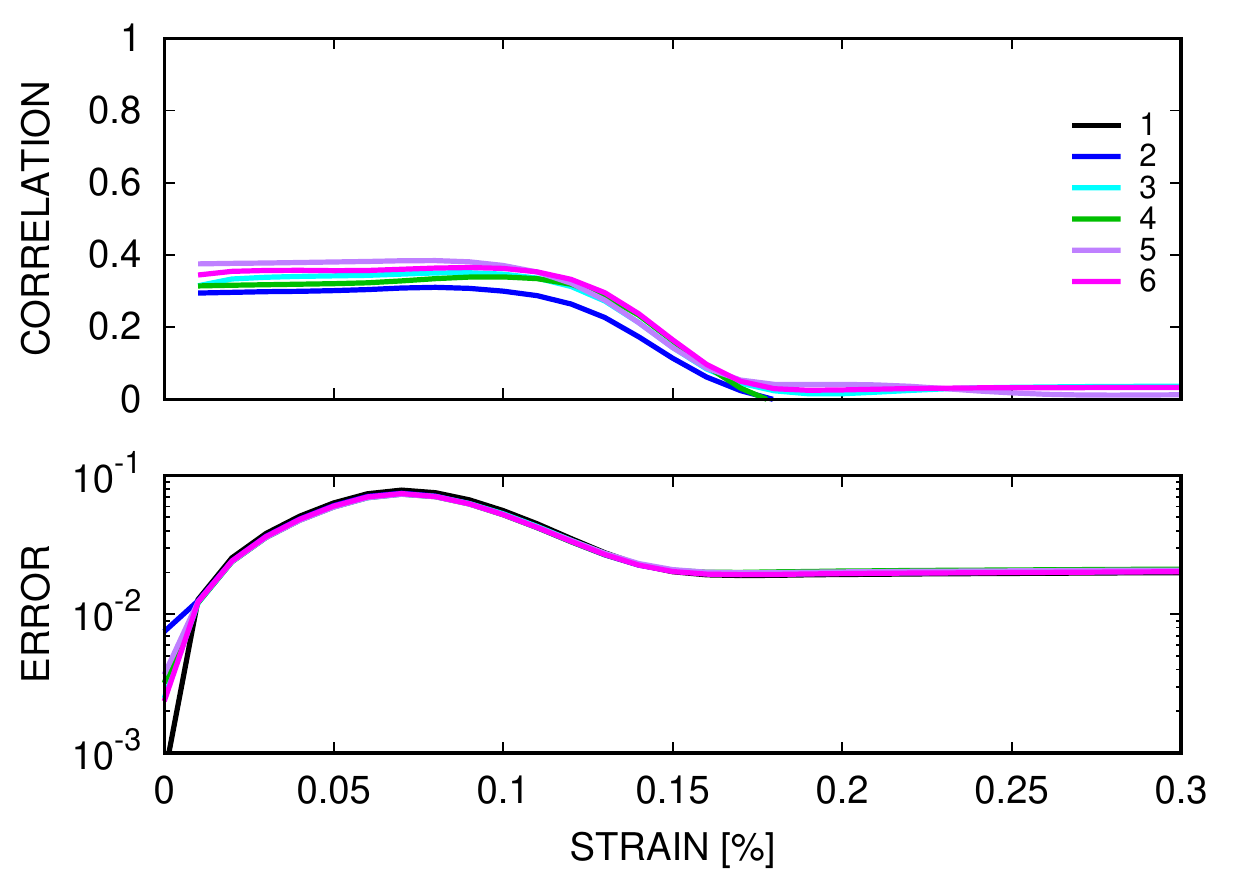}}
{\includegraphics[width=0.45\textwidth]{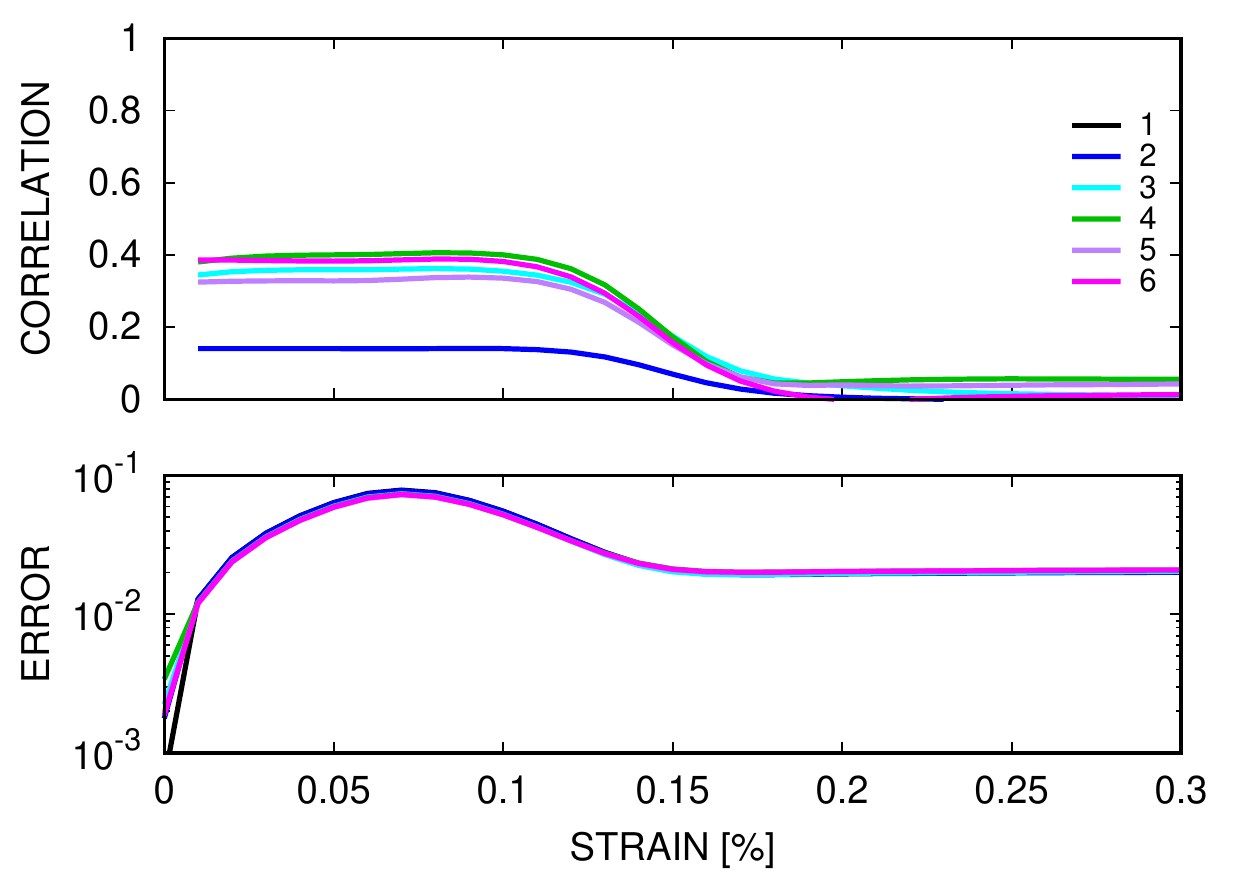}}
\caption{\RGCNN}
\end{subfigure}
\caption{Architecture comparison (2D CP data): (a) CNN, (b) \DGCNN: GCN applied directly grid, (c) \RGCNN: GCN applied to clustered/segmented data.
Left panels: 1 convolutional layer, 1 dense layer;
right panels: 2 convolutional layers, 1 dense layer.
}
\label{fig:architecture_comparison}
\end{figure}

\fref{fig:architecture_comparison} shows the correlations $\corr(t)$ and errors $\error(t)$ over time for the three architectures for a range of filters $\Nf \in \{1,\ldots,6\}$ for 1 or 2 convolutional layers ($\Nc =\{1,2\}$) and one dense layer ($\Nd=1$).
Note that two dense layers, $\Nd=2$, produced similar results.
Generally the correlation of all three hybrid CNN-RNNs is better earlier in the process, while the error tends to peak early on near the end of the elastic regime where the response variance is highest.
Referring to \fref{fig:stress_history}a, we observe that elastic-to-plastic transition occurs around 0.1\% and this transition is apparent in \fref{fig:architecture_comparison} where the correlation appears to transition between two plateaus.
All architectures improve with more filters although there is a clearly a limit to the improvement, which suggests a small number of relevant features.
It is also apparent that more filters are needed to capture the later plastic regime accurately than the initial elastic regime.

The simplest CNN models (fewest parameters) that have the best performance are: $\arch{4}{1}{1}$ with 269 parameters, $\arch{3}{2}{1}$ with 271 parameters  (shown in \fref{fig:architecture_comparison}), and $\arch{3}{1}{2}$ with 187 parameters, $\arch{2}{2}{2}$ with 151  (not shown).
This demonstrates the fungibility of the nodes in the network.
The proposed \DGCNN~with $\Nc=1$ with either $\Nd=\{1,2\}$ does well in the elastic regime for $\Nf > 3$ but achieves no greater than $0.7$ correlation for plastic for all cases with $\Nf < 9$; however, with a second convolutional layer three filters are sufficient to achieve the accuracy of the best CNN architectures.
This indicates that second nearest neighbors are required to represent the plastic flow well using the standard GCN filter in the \DGCNN~architecture.
This finding will be revisited in \sref{sec:filter}.
Lastly, all the variants of \RGCNN~with angle and volume node features gave similar (poor) performance.
The \RGCNN~clearly require more features for improvement.
This finding will be expanded on in \sref{sec:features}.

\tref{tab:parameter_counts} compares the number of trainable parameters for the three types of convolutional networks.
Clearly, the additional weights in the pixel-based convolutional layers incur a cost in complexity and training.
Also the parameter complexity of the \DGCNN~is essentially equivalent to the \RGCNN~since the same filters are being used on different graphs and data.
The data is certainly different, with the \RGCNN~storing more features on a more compact adjacency than the \DGCNN; with sparse storage this is not a significant advantage for moderately sized meshes.

\begin{table}
\centering
\begin{tabular}{|c|c|c|c|}
\hline
& CNN & \DGCNN & \RGCNN \\
\hline
$\arch{1}{1}{1}$ & 41 & 26 & 27 \\
$\arch{2}{1}{1}$ & 99 & 75 & 71 \\
$\arch{4}{1}{1}$ & 269 & 209 & 213 \\
$\arch{6}{1}{1}$ & 511 & 421 & 427 \\
\hline
$\arch{1}{2}{1}$ & 55 & 32 & 33 \\
$\arch{2}{2}{1}$ & 145 & 69 & 85 \\
$\arch{4}{2}{1}$ & 433 & 245 & 249 \\
$\arch{6}{2}{1}$ & 865 & 487 & 493 \\
\hline
\end{tabular}
\caption{Parameter counts for architectures with GCN filters applied to 2D CP data}
\label{tab:parameter_counts}
\end{table}

\subsection{Comparison of convolutional filters} \label{sec:filter}

\begin{figure}
\centering
\begin{subfigure}[c]{0.55\textwidth}
{\includegraphics[width=1.0\textwidth]{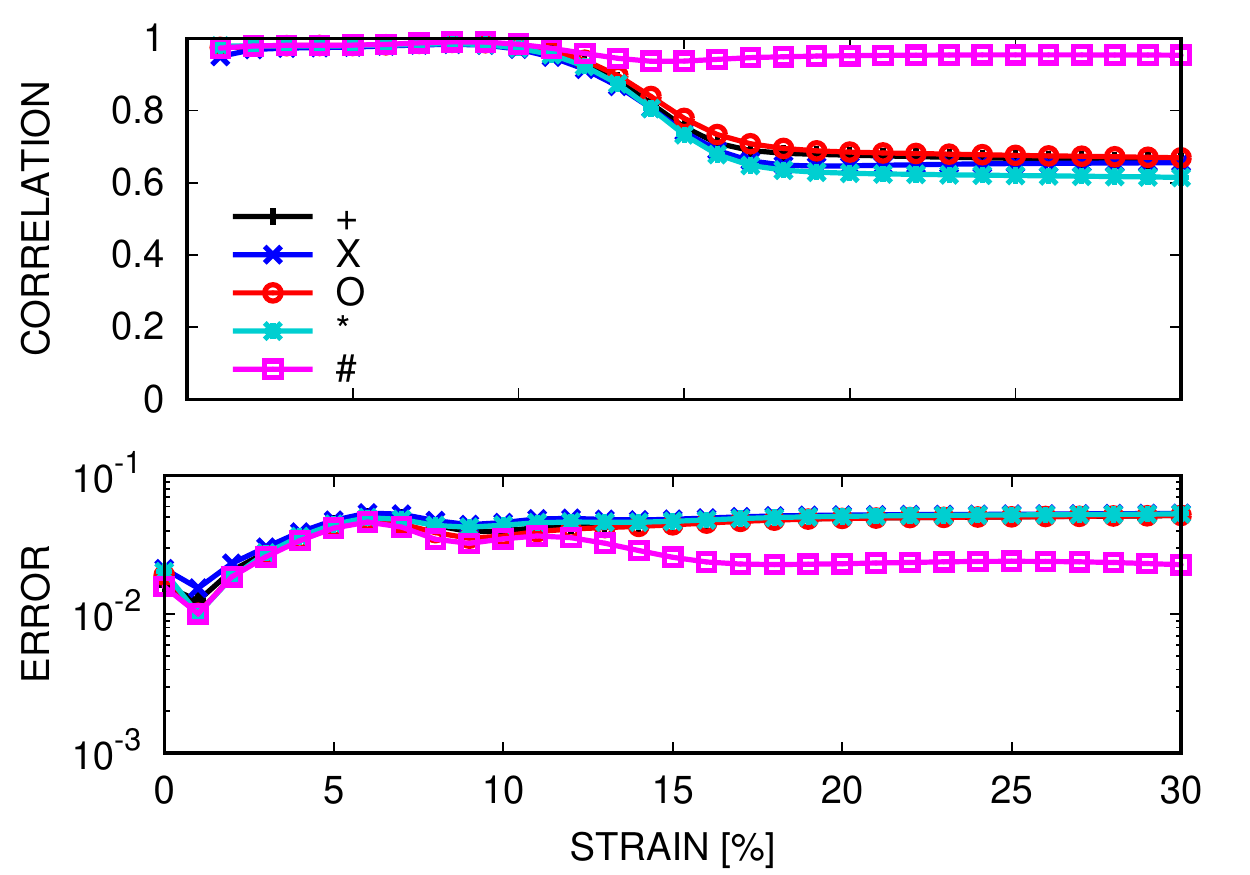}}
\caption{}
\end{subfigure}
\begin{subfigure}[c]{0.55\textwidth}
{\includegraphics[width=1.0\textwidth]{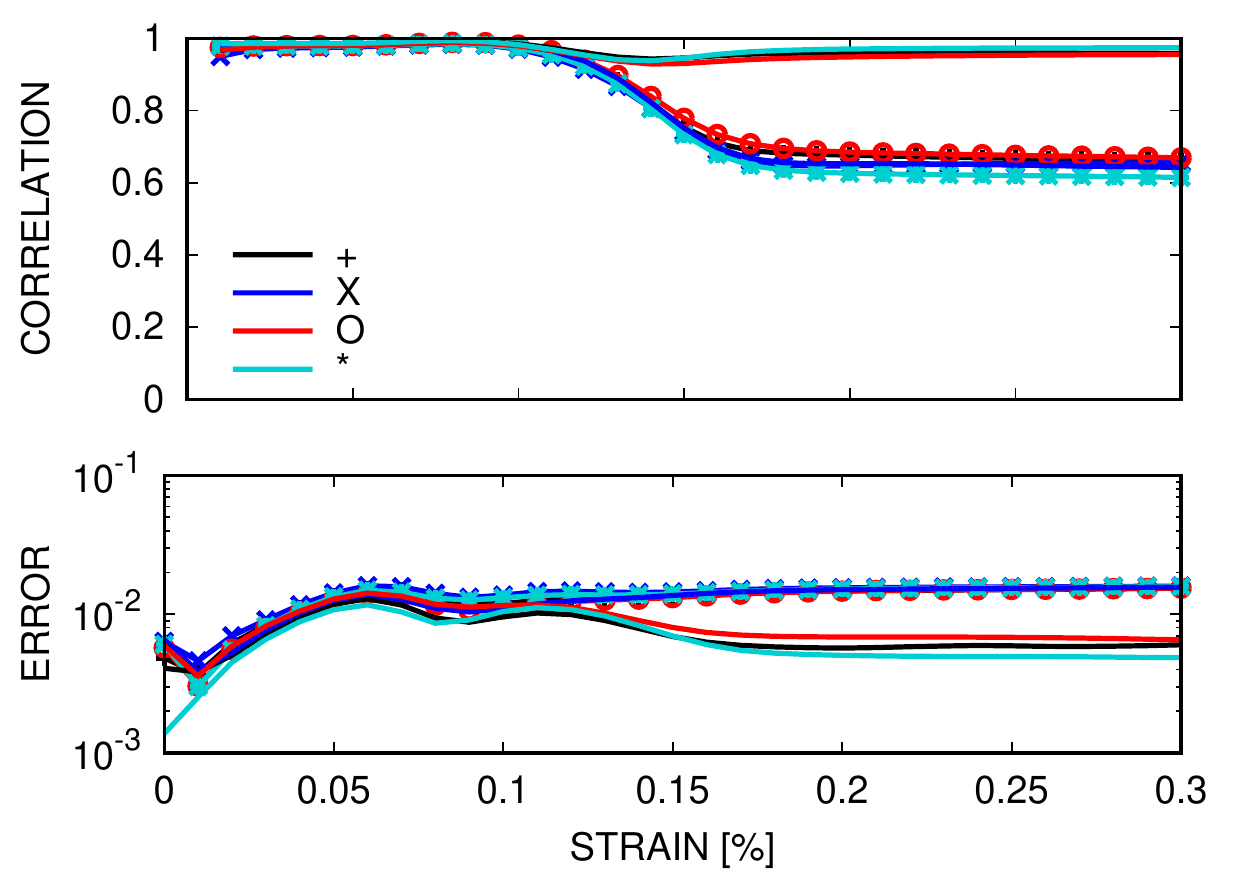}}
\caption{}
\end{subfigure}
\caption{Filter comparison (2D CP data):
(a) patterns illustrated \fref{fig:filters} and
(b) edge,vertex, and both filters augmented with an independent center weight.
}
\label{fig:filter_comparison}
\end{figure}

Motivated by the fact that a CNN can achieve good performance with only one convolutional layer we tried richer variants of the standard GCN filter (where the self-weight is set equal to the neighbor weight).
Using the patterns shown in \fref{fig:filters}, we explored their relative performance with one convolutional layer ($\Nc=1$) and four filters ($\Nf=4$).
\fref{fig:filter_comparison}a shows that patterns +, X, O, which only have one trainable weight and have interactions with edge, vertex-only, and edge+vertex neighbors, respectively, have comparable and less than satisfactory performance.
Even the $\ast$ pattern, where edge and node neighbors are given separate weights (2 independent weights) has an inferior performance to the CNN (which has 9 independent weights).
Only the \# pattern, where the center pixel is given an independent weight from the vertex neighbors, has performance on par with the CNN.
Given this finding we endowed each of the basic patterns \{+, X, O, $\ast$\} with an independent central weight.
As shown in \fref{fig:filter_comparison}b, this was sufficient to improve the performance of all but the X (vertex only neighbors) to be comparable with the CNN.
It is physically plausible that that edge neighbors have a stronger influence than vertex neighbors and it appears that they are required for the filters to learn predictive interactions.

\subsection{Selected features} \label{sec:features}

\begin{figure}
\centering
{\includegraphics[width=0.55\textwidth]{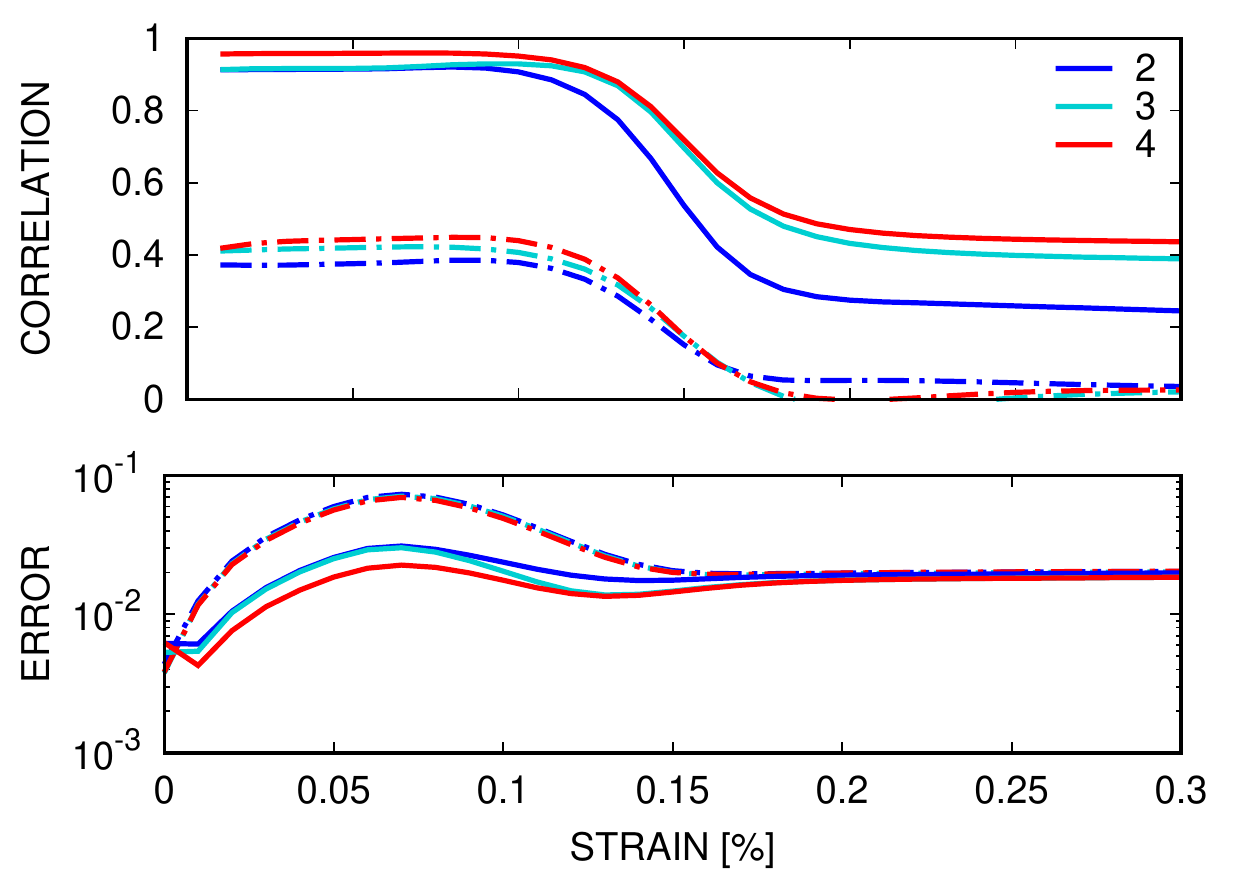}}
\caption{Comparison of \RGCNN~with increasing number of features (2D CP data).
Features: angle, volume, area, surface area.
Graph labels correspond to
2: \{angle, volume\},
3: \{angle, volume, area\}, and
4: \{angle, volume, area, surface area\}.
Dashed lines: self-weight equal to neighbor weight; solid lines: self-weight independent of neighbor weight.
}
\label{fig:feature_comparison}
\end{figure}

In \sref{sec:components} we observed that the feature-dependent \RGCNN~formulation had sub-par performance especially in the plastic regime that was not improved with a more complex GCNN component.
\fref{fig:feature_comparison} shows the performance of a $\arch{4}{2}{1}$ \RGCNN~network does improve with an expanded feature set.
Here, in addition to the orientation angle and the volume associated with the clusters/grains represented by the graph nodes (2 total features), we added the surface area of each grain (3 total features) and the area of the grain that is on the surface of the cell (4 total features) to the node features.
The improvement with these additional features is marginal.
However, if we allow for an independent self-weight by changing the standard GCN pattern to the \# pattern, as in the previous section, the performance is dramatically improved and the improvement with additional features is more distinct.
Although these networks have considerably smaller adjacency matrices than \DGCNN due to the clustering of the pixels, the performance is sub-par, particularly in the plastic regime.
This serves as an illustration of the difficulty of improvement by feature selection, as opposed to deep learning.%
\footnote{Note only elastic response was modeled in \cref{vlassis2020geometric}.}

\subsection{Data efficiency} \label{sec:efficiency}

\begin{figure}
\centering
{\includegraphics[width=0.55\textwidth] {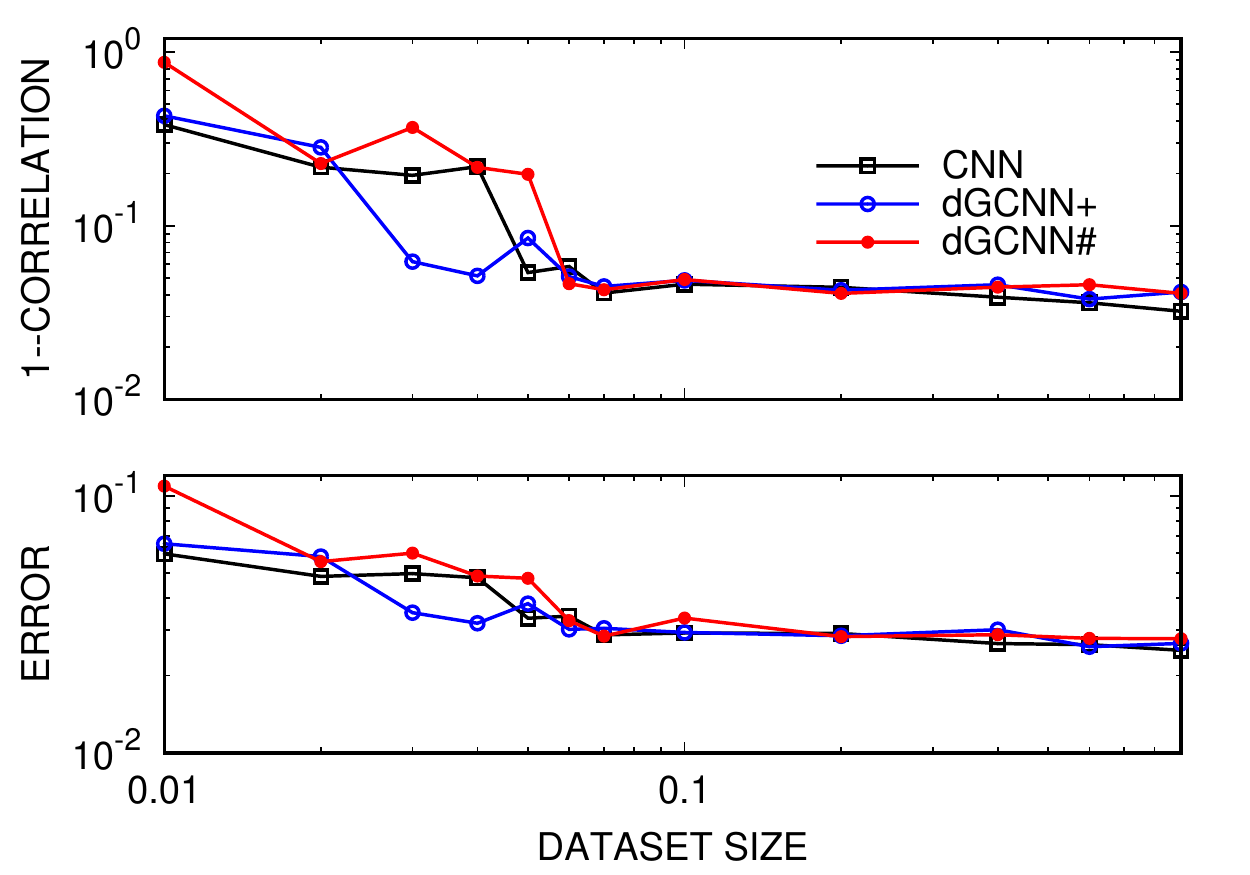}}
\caption{2D CP data: data efficiency comparison for $\arch{4}{1}{1}$ CNN with 269 parameters, $\arch{4}{2}{1}$ dGCNN+ 317 parameters, and $\arch{4}{1}{1}$ dGCNN\# with 249 parameters.
Note $1-C(t)$ is being plotted in the upper panel.
}
\label{fig:datasize_comparison}
\end{figure}

As a last trial with the 2D CP dataset, we investigated how efficient the best networks are with smaller datasets.
Here we compared (a) $\arch{4}{1}{1}$ CNN with 269 parameters, (b) $\arch{4}{2}{1}$ \DGCNN~with the + pattern (GCNN+) and 317 parameters, and (c) $\arch{4}{1}{1}$ \DGCNN~with the \# pattern (GCNN\#) and 249 parameters.
The training set was reduced from 80\% of the 12,000 realizations (9600) by a fraction that ranged from 0.01 to 1.0.
The test set was a fixed 20\% (2400) of the full realizations and the results where averaged over 9 trials.
\fref{fig:datasize_comparison} shows that the majority of learning (improvement in accuracy) occurs by the time the training size is approximately equal to the number of parameters.
After the step-down in error (at 0.04 of the total training set for the CNN, at 0.02 for the GCNN+, and at 0.05 GCNN\#) the improvement is relatively slow but steady.
This data demonstrates that these small networks can be effective at the prediction of the homogenized response task, \eref{eq:problem}, with much smaller data sets.

\subsection{Boosting with preconceived features} \label{sec:boost}

\begin{figure}
\centering
{\includegraphics[width=0.55\textwidth] {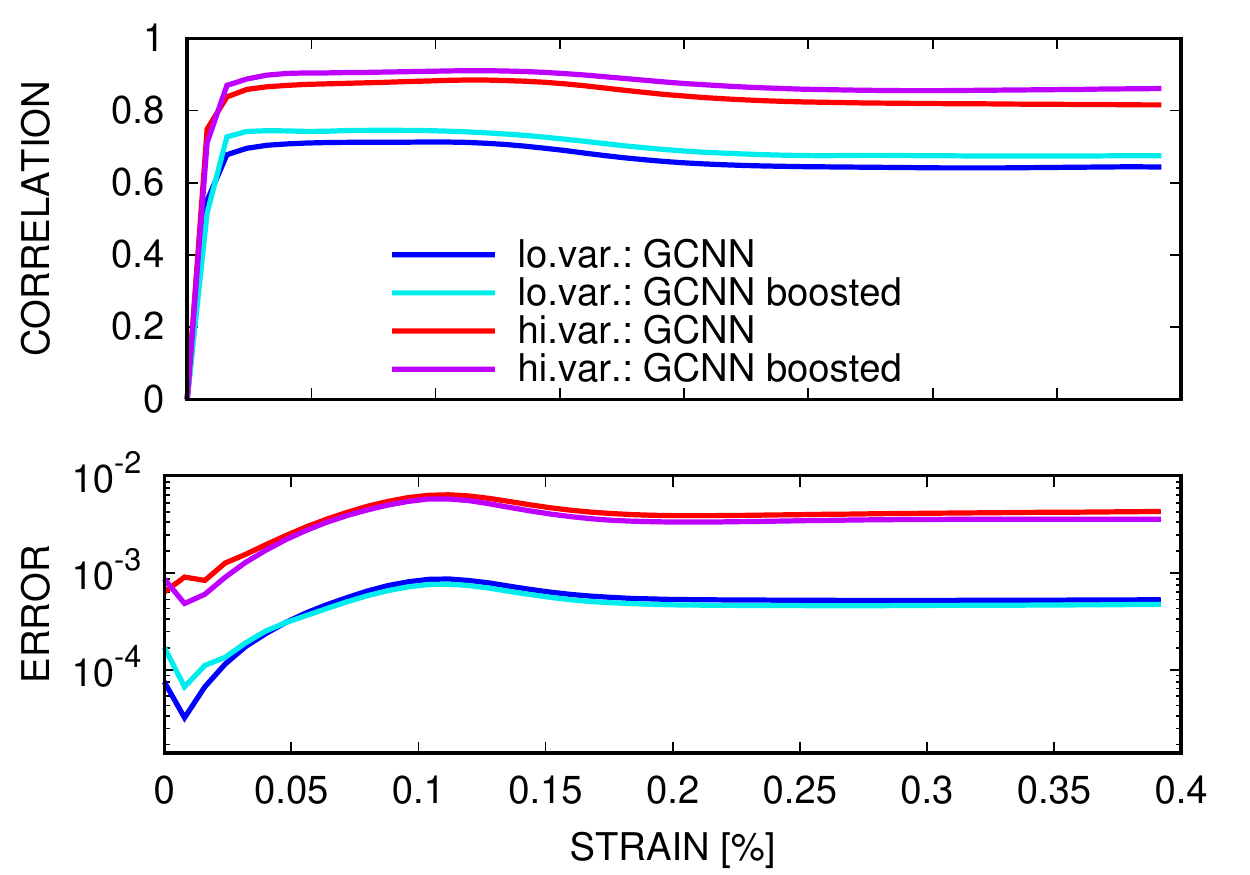}}
\caption{The effect of boosting the microstructural input field with with volume fraction.
}
\label{fig:boost_comparison}
\end{figure}

Now that we have discovered effective adjacencies and guidance on architecture hyper-parameters, we turn to using the 3D CP datasets.
Motivated by the fact that some features of the image $\phib(\Xb)$ have obvious bearing on the output due to physical reasoning, we boosted the \DGCNN~with some of the features we employed with the solely feature-based \RGCNN.
The proposed architecture can accommodate pre-selected features by simply augmenting the image/cell microstructural field $\phib$ with additional channels.
\fref{fig:boost_comparison} shows the effect of adding a channel with the volume fraction of the associated grain to each pixel.
Clearly there is a distinct and uniform benefit; however, it is somewhat marginal due to the fact that the selected feature is likely, at least partially, redundant/correlated with the output of the trainable filters.
Additional means of augmenting with more global data, such as the average grain size or equivalently the grain density, via inputs concatenated to the pooling layer (gray in \fref{fig:architecture}) output going  to the dense layers (green in \fref{fig:architecture}) would also likely prove beneficial.

\subsection{Generalizability} \label{sec:generalizability}

\begin{figure}
\centering
\begin{subfigure}[c]{0.55\textwidth}
{\includegraphics[width=0.9\textwidth]{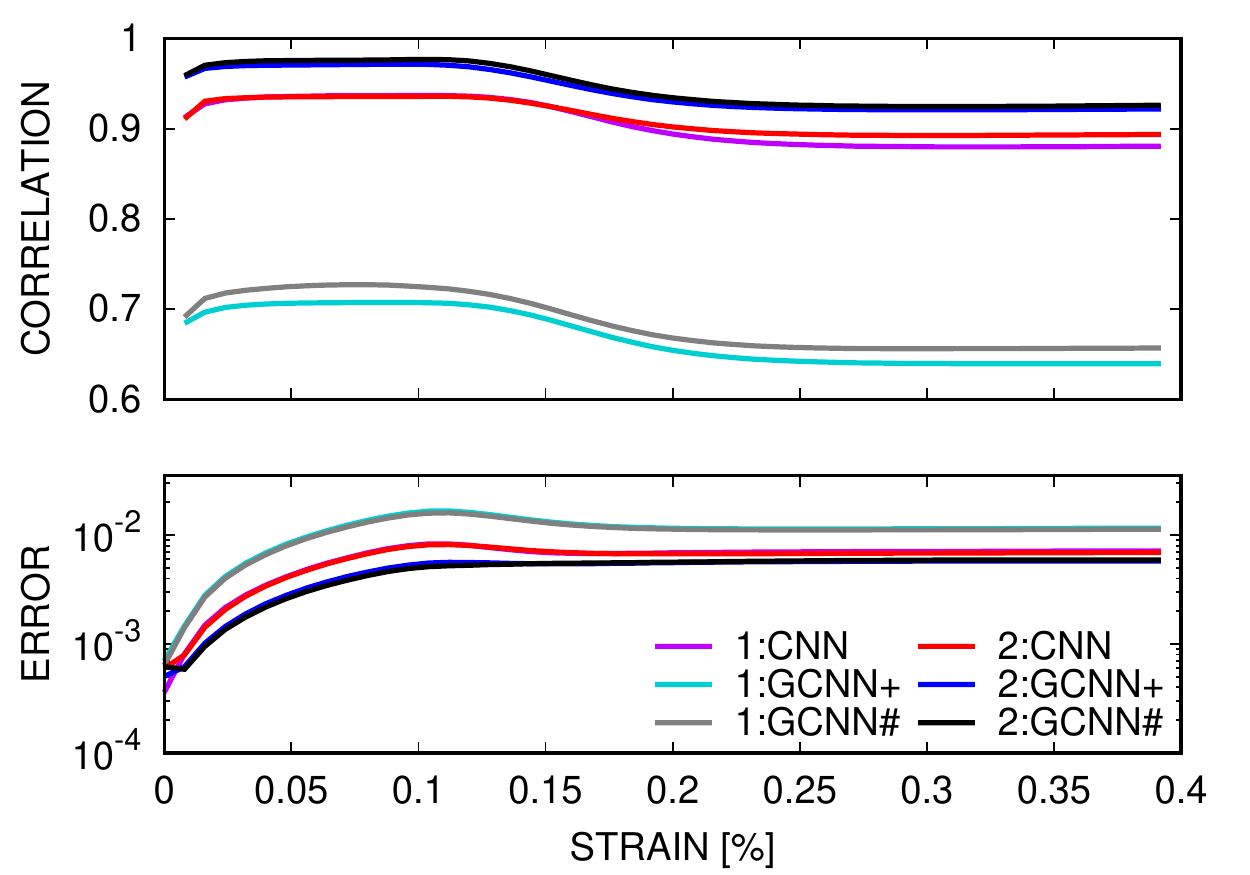}}
\caption{low variance}
\end{subfigure}
\begin{subfigure}[c]{0.55\textwidth}
{\includegraphics[width=0.9\textwidth]{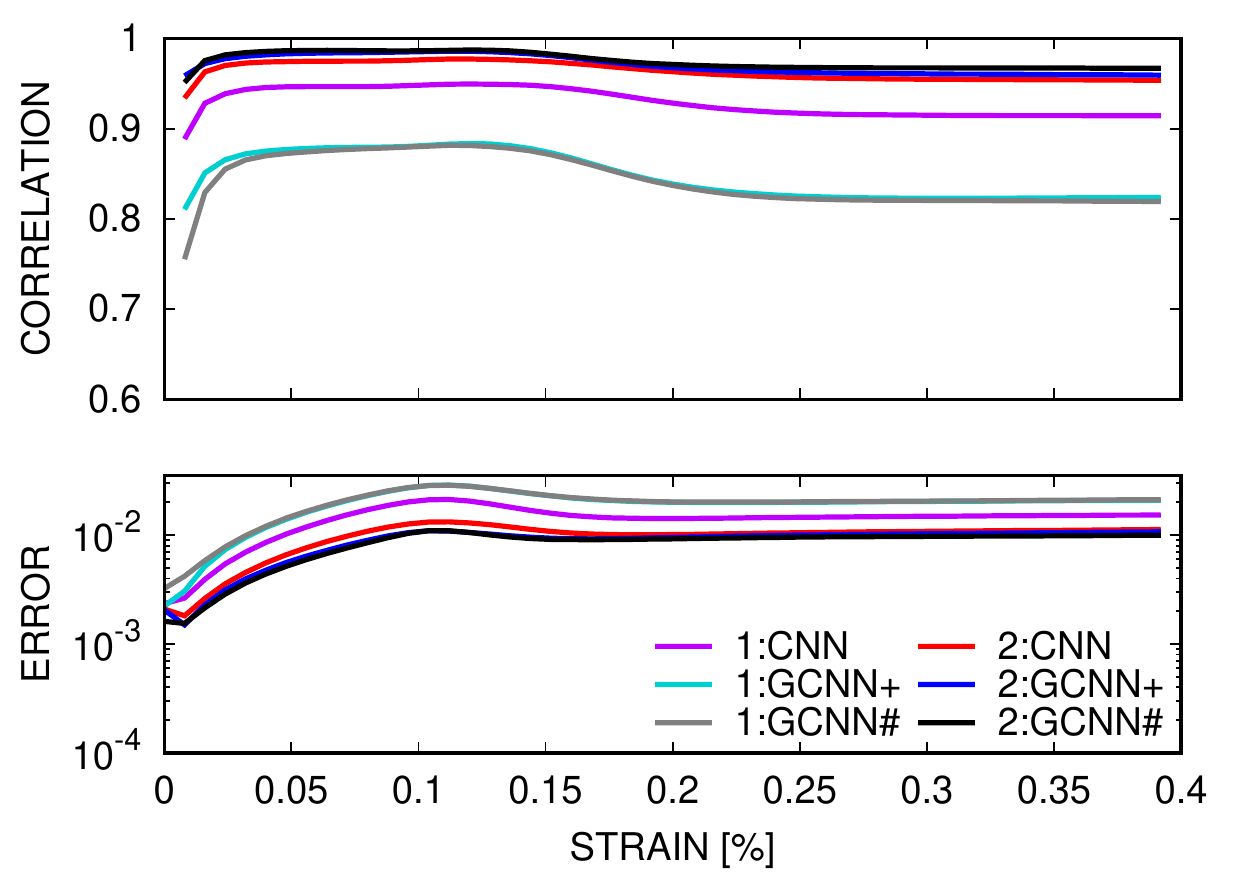}}
\caption{high variance}
\end{subfigure}
\caption{3D data: performance of $\arch{32}{1}{1}$ and $\arch{32}{2}{1}$ configurations of a CNN, \DGCNN{+}, and \DGCNN\# for the low and high variance datasets
(2:GCNN+ denotes a $\arch{32}{2}{1}$ direct graph CNN using a ``+'' pattern).
}

\label{fig:data_variance}
\end{figure}

Using CNN, GCNN{+}, and GCNN{\#} with 32 filters, 1 or 2 convolutional layers, and 1 dense, layer \fref{fig:data_variance} shows that the proposed architecture with the \# filter with an independent central weight can outperform a corresponding CNN and GCNN{+}.
The benefits of the more complex $\arch{32}{2}{1}$ configuration over the $\arch{32}{1}{1}$ appear to be most significant for the two \DGCNN{s}.
It is also apparent that all the types of convolutional neural networks perform better, at least in terms of correlation, on the higher variance dataset than on the lower variance dataset.

\begin{figure}
\centering
\begin{subfigure}[c]{0.55\textwidth}
{\includegraphics[width=0.9\textwidth]{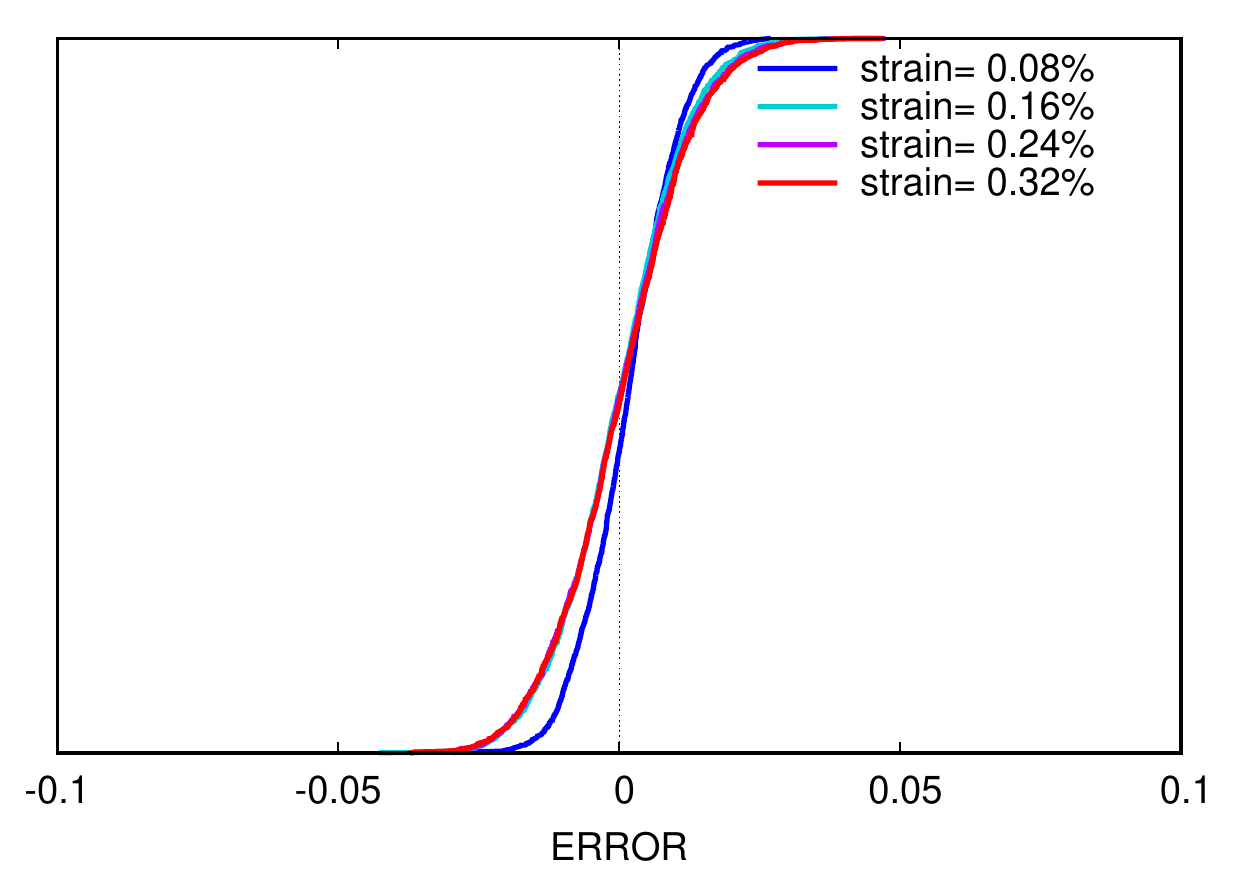}}
\caption{low variance}
\end{subfigure}
\begin{subfigure}[c]{0.55\textwidth}
{\includegraphics[width=0.9\textwidth]{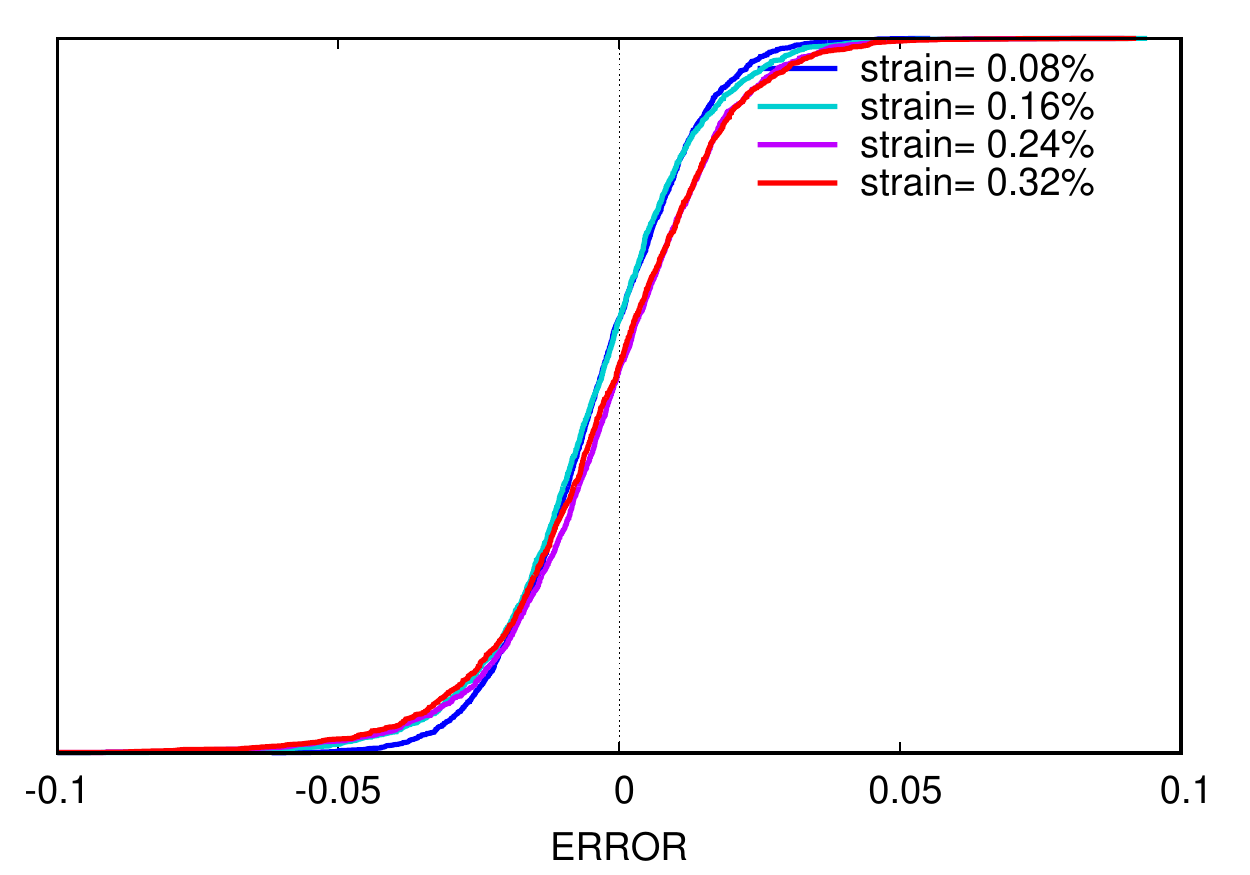}}
\caption{high variance}
\end{subfigure}
\caption{3D data: cumulative distribution of error for various strains ($\arch{32}{2}{1}$ \DGCNN\#).
}
\label{fig:error_CDFs}
\end{figure}

\begin{figure}
\centering
\begin{subfigure}[c]{0.55\textwidth}
{\includegraphics[width=0.9\textwidth]{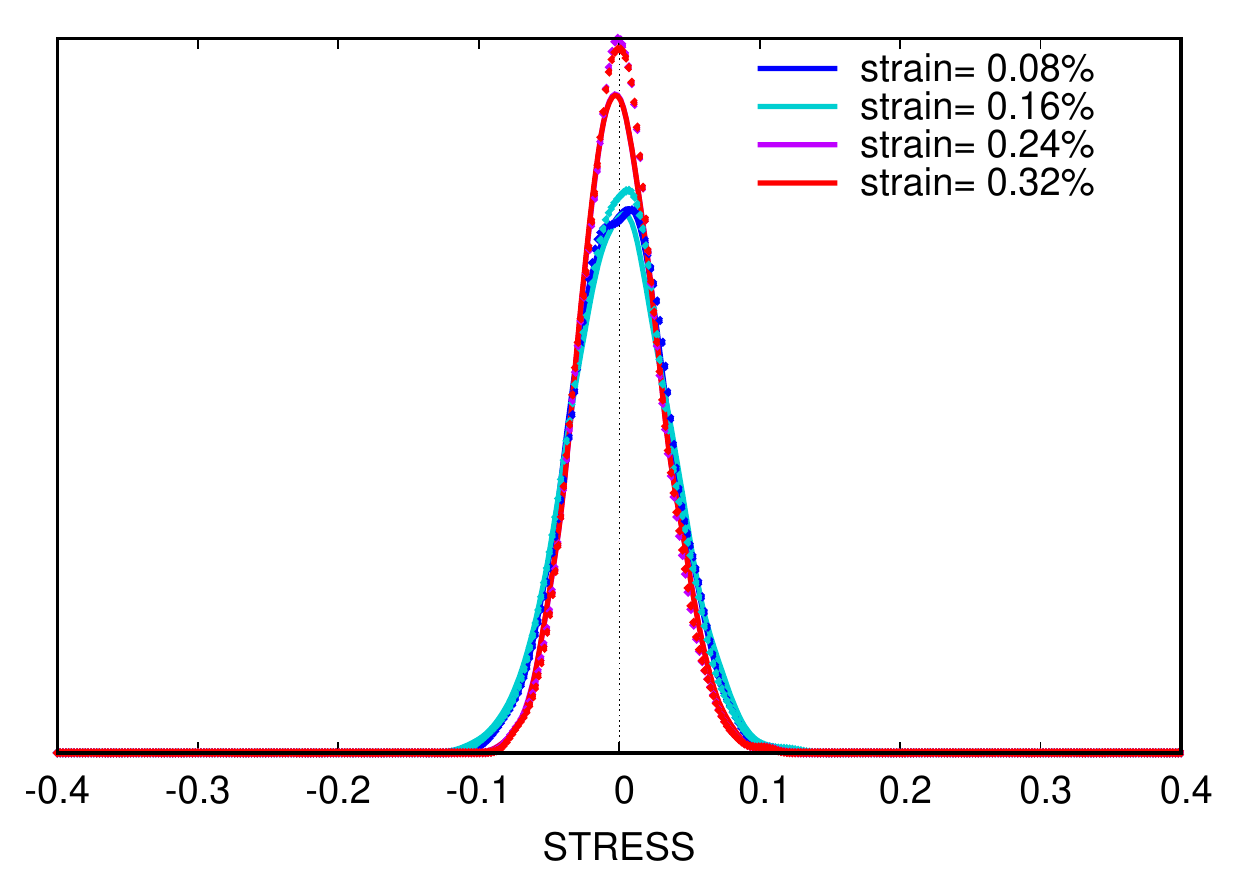}}
\caption{low variance}
\end{subfigure}
\begin{subfigure}[c]{0.55\textwidth}
{\includegraphics[width=0.9\textwidth]{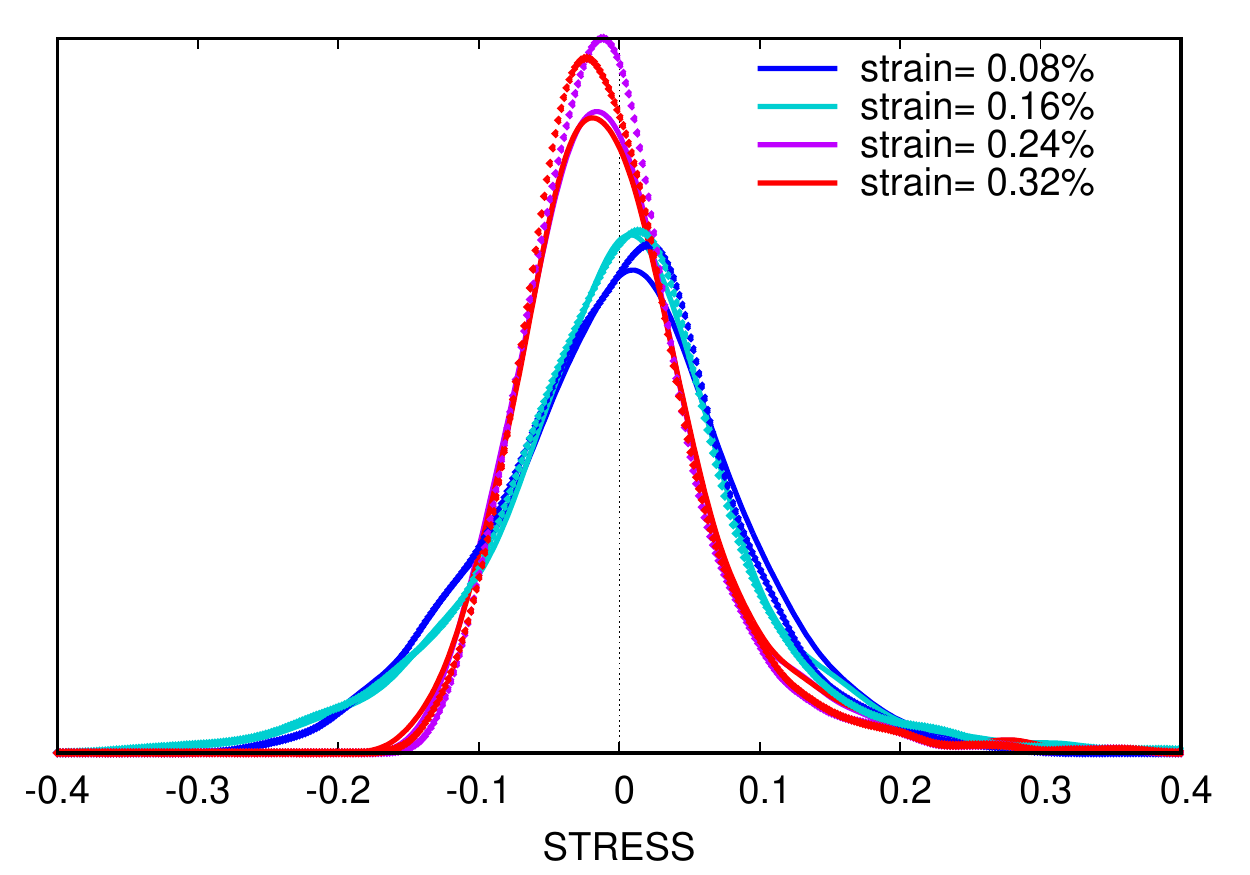}}
\caption{high variance}
\end{subfigure}
\caption{3D data: distribution of true and predicted stress values for various strains ($\arch{32}{2}{1}$ \DGCNN\#).
}
\label{fig:pred_PDFs}
\end{figure}

Now focusing on the GCNN{\#}, \fref{fig:error_CDFs} shows the distribution of RMSE errors (\eref{eq:error}) is approximately Gaussian with some outlier errors above 5\% for the high variance dataset and 3\% for the low variance dataset.
A direct comparison of the true and predicted values over a sequence of strains, shown in \fref{fig:pred_PDFs}, indicates that the GCNN{\#} overpredicts values near the mean which may be due it being harder to distinguish near-mean response microstructures from those that produce extreme/outlier responses.

\begin{figure}
\centering
{\includegraphics[width=0.55\textwidth]{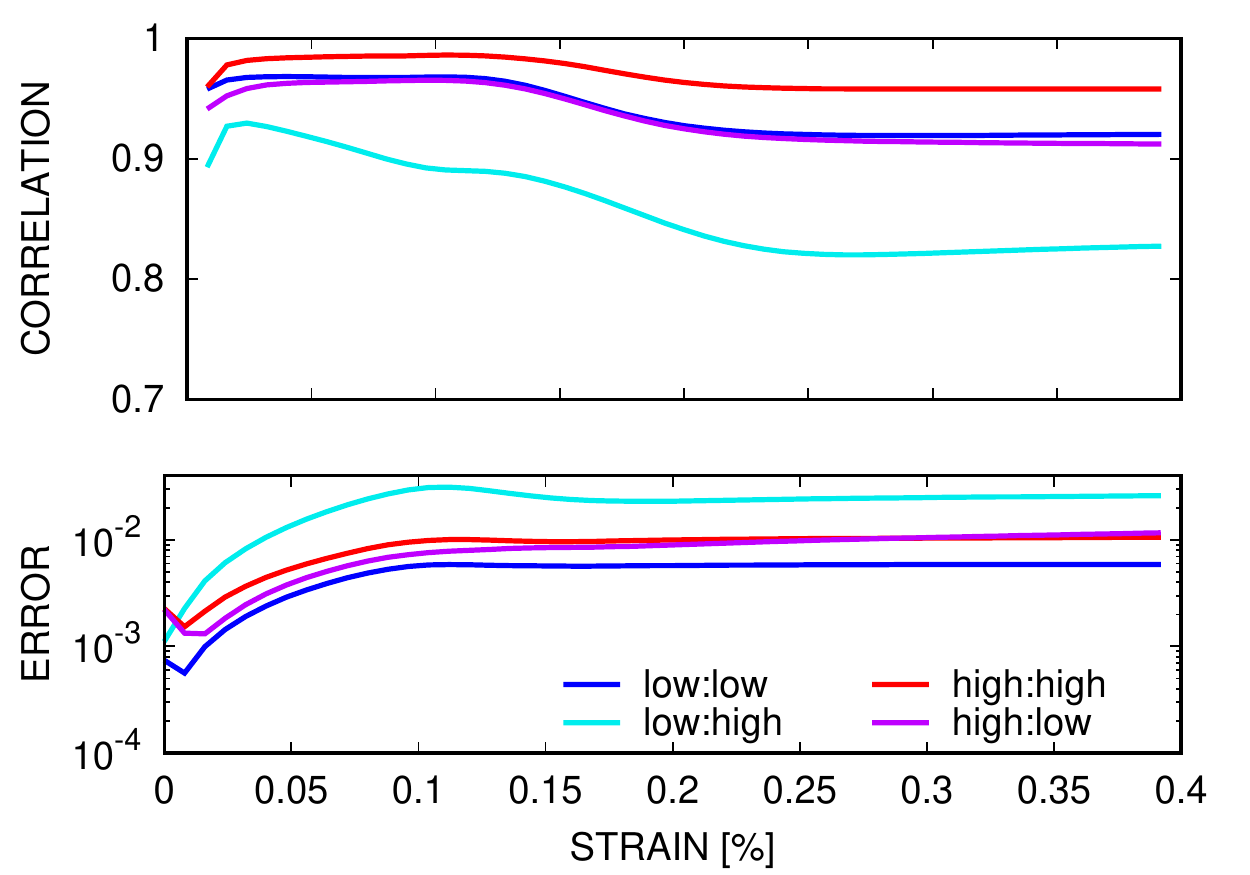}}
\caption{3D data: generalizability results for train:test pairs ($\arch{32}{2}{1}$ \DGCNN\#).
}
\label{fig:generalizability}
\end{figure}

Following this conjecture, \fref{fig:generalizability} illustrates that training on the low variance ensemble and testing on the high variance ensemble (which also has a different mean) does relatively poorly compared to the reverse.
It appears that the network generalizes well to different distributions of inputs if they are in the span of the training set \ie it does well at interpolation and less well at extrapolation to potential out-of-distribution samples.

\section{Discussion}\label{sec:discussion}

The response of microstructures for use in multiscale simulations, structure-property investigations, and uncertainty quantification can be accurately modeled with graphs.
The proposed formulation used the topology of the data discretization directly instead of a segmentation or clustering of the image data.
This aspect should have particular advantages for image data where the segmentation is not obvious, hard to compute, or is obscured by noise.
Furthermore, it has a simple implementation and avoids the need for feature engineering, but can benefit from it.
The architecture draws on both purely graph-based networks and permutationally invariant convolutional filters.
We demonstrated that endowing the widely-used GCN filter \cite{kipf2016semi} with an independent self-weight (as suggested by the reduction of the ChebNet) can significantly improve accuracy without adding additional layers and their parameters.
The independent self-weight allows for differencing the node data of the self and its neighbors instead of only averaging.
This can be seen as giving the filter the ability to infer edge features between the central pixel and its neighbors.
For physical problems driven by gradients this change to the filter is important.
We also found that pixel edge neighbors are more crucial for a predictive model than vertex-only neighbors.
Lastly we were able to demonstrate that small, efficient graph convolutional networks can be effective at the task of predicting the homogenized evolution of complex microstructure.
This has significant applications in sub-grid constitutive models in large scale simulations, structure-property property investigations, and material uncertainty quantification.

An apparent downside of the proposed approach is the graph and its adjacency grows with resolution of image (number of pixels/elements).
This issue is partially offset by sparse storage of the adjacency matrix, in general, and largely ameliorated by data that is on the same discretization.
In future work we will investigate low-rank approximations to the adjacency matrix \cite{savas2011clustered,richard2012estimation,lebedev2014speeding,tai2015convolutional,kanada2018low}, dimensionality reduction techniques \cite{belkin2003laplacian,he2004locality}, and the use of graph auto-encoders \cite{kipf2016variational,liao2016graph,hasanzadeh2019semi,salehi2019graph} to reduce the mesh-based graphs in-line.
We are also pursuing the larger topic of processing image with multi-resolution filters \cite{zhang2018multiresolution}, \eg spanning the pixel to the cluster level.

\section*{Acknowledgments}
This material is based upon work supported by the U.S. Department of Energy, Office of Science, Advanced Scientific Computing Research program. Sandia National Laboratories is a multimission laboratory managed and operated by National Technology and Engineering Solutions of Sandia, LLC., a wholly owned subsidiary of Honeywell International, Inc., for the U.S. Department of Energy's National Nuclear Security Administration under contract DE-NA-0003525. This paper describes objective technical results and analysis. Any subjective views or opinions that might be expressed in the paper do not necessarily represent the views of the U.S. Department of Energy or the United States Government.


\appendix
\section{The Graph Convolutional Network} \label{app:gcn}
As mentioned in the Introduction, the Kipf and Welling Graph Convolutional Network (GCN) \cite{kipf2016semi} provides an innovative, expressive graph convolutional network built in sequentially applied layers with local action.
Here we give a brief synopsis of their development.

The objective is to efficiently apply a graph filter $g_\thetab = \diag(\thetab)$ where the parameter vector $\thetab$ has $N_\text{nodes}$ entries.
Convolution of the filter $g_\thetab$ and data $\xs$ on a graph can be expressed as \cite{hammond2011wavelets}
\begin{equation} \label{eq:graph_convolution}
g_\thetab \ast \xs =  \Us g_\thetab \ \Us^T \xs
\end{equation}
analogous to the classical convolution theorem.
This formulation is connected to the (normalized) graph Laplacian $\Ls$ and its spectral representation
\begin{equation}
\Ls = \Is - \Ds^{-1/2} \As \Ds^{-1/2} = \Us \Lambdab \Us^T
\end{equation}
where $\As$ is the binary adjacency matrix, $\Ds$ is the associated degree matrix,  $\Us$ is the matrix of eigenvectors and $\Lambdab$ is the diagonal matrix of eigenvalues.
The formulation for graph convolution in \eref{eq:graph_convolution}, in turn can be approximated by an expansion of Chebyshev polynomials $T_k$ \cite{defferrard2016convolutional}
\begin{equation} \label{eq:cheby}
g_\thetab \ast \xs = \left( \Us g_\thetab  \Us^T \right) \xs
\approx \sum_k \vartheta_{k=0}^K T_k(\tilde{\Ls}) \, \xs
\end{equation}
where $\tilde{\Ls} = \frac{2}{\lambda_\text{max}} \Ls - \Is$ and $\lambda_\text{max}$ is the maximum eigenvalue of $\Ls$.

To this approximation Kipf and Welling make a number of additional simplifications.
First they approximate the maximum eigenvalue $\lambda_\text{max} \approx 2$ so that $\tilde{\Ls} = \Ls - \Is$ \ie $\tilde{\Ls}$ is the graph Laplacian with added self-loops/interactions.
Next, they truncate the expansion in \eref{eq:cheby} at $K=1$ so that
\begin{equation}
g_\thetab \ast \xs
\approx \vartheta_0 \Is \xs - \vartheta_1 \Ds^{-1/2} \As \Ds^{-1/2} \xs
\end{equation}
This effectively reduces the number of free parameters in the filter from $N_\text{nodes}$ to 2.
This has the tremendous advantage of cheap and local action.
The expressiveness of a network built on these layers is controllable by the GCNN depth (number of layers).
Lastly they further collapse the number of free parameters from 2 to 1 by setting $\vartheta_1 = - \vartheta_0$

\end{document}